\documentclass[letterpaper, 10 pt, conference]{ieeeconf}  

\IEEEoverridecommandlockouts                              

\usepackage{amsfonts}
\usepackage{graphicx}
\usepackage{subfig}
\usepackage{rotating}
\graphicspath{{figures/}}
\usepackage{multirow}
\usepackage{hhline}
\usepackage{amsmath}
\usepackage{adjustbox}
\usepackage{caption}
\usepackage{fancyhdr}
\usepackage{rotating}
 \usepackage[table,xcdraw]{xcolor}

\usepackage{calrsfs}
\DeclareMathAlphabet{\pazocal}{OMS}{zplm}{m}{n}

\newcommand{\RNum}[1]{\lowercase\expandafter{\romannumeral #1\relax}}
\usepackage{array}
\newcolumntype{C}[1]{>{\centering\arraybackslash}p{#1}}
\renewcommand{\arraystretch}{1.3}
\usepackage{multirow}
\makeatletter
\newcommand{\removelatexerror}{\let\@latex@error\@gobble}
\makeatother
\usepackage[linesnumbered,ruled,vlined]{algorithm2e}
\SetKwInput{KwInput}{Input}                
\SetKwInput{KwOutput}{Output}              
\SetKwInput{KwConfig}{Configuration}              
\usepackage{bm} 
\usepackage{booktabs}

\usepackage{subfig}
\usepackage{mwe}

\newlength{\tempheight}
\newlength{\tempwidth}

\newcommand{\rowname}[1]
{\rotatebox{90}{\makebox[\tempheight][c]{\textbf{#1}}}}

\newcommand{\columnname}[1]
{\makebox[\tempwidth][c]{\textbf{#1}}}

\title
{\LARGE \bf
Neuromorphic Eye-in-Hand Visual Servoing}

\author{Rajkumar~Muthusamy$^{1}$,~Abdulla~Ayyad$^{1}$, ~Mohamad ~Halwani$^{1}$, ~Yahya~Zweiri$^{1,2}$, ~Dongming~Gan$^{3}$, \\ and Lakmal Seneviratne$^{1}$
\thanks{$^{1}$  Khalifa University Center for Autonomous Robotic Systems (KUCARS), Khalifa University of Science and Technology, Abu Dhabi, UAE. Email: {\tt\small \{rajkumar.muthusamy@ku.ac.ae\}} }
\thanks{$^{2}$ Faculty of Science, Engineering and computing, Kingston University, London SW15 3DW, UK. }
\thanks{$^{3}$ School of Engineering Technology, Purdue University, West Lafayette, IN 47907, USA. }
}

\begin{document}

\maketitle
\thispagestyle{empty}
\pagestyle{empty}

\begin{abstract}

Robotic vision plays a major role in factory automation to service robot applications. However, the traditional use of frame-based camera sets a limitation on continuous visual feedback due to their low sampling rate and redundant data in real-time image processing, especially in the case of high-speed tasks. Event cameras give human-like vision capabilities such as observing the dynamic changes asynchronously at a high temporal resolution ($1\mu s$) with low latency and wide dynamic range. 

In this paper, we present a visual servoing method using an event camera and a switching control strategy to explore, reach and grasp to achieve a manipulation task. We devise three surface layers of active events to directly process stream of events from relative motion. A purely event based approach is adopted to extract corner features, localize them robustly using heat maps and generate virtual features for tracking and alignment. Based on the visual feedback, the motion of the robot is controlled to make the temporal upcoming event features converge to the desired event in spatio-temporal space. The controller switches its strategy based on the sequence of operation to establish a stable grasp. The event based visual servoing (EVBS) method is validated experimentally using a commercial robot manipulator in an eye-in-hand configuration. Experiments prove the effectiveness of the EBVS method to track and grasp objects of different shapes without the need for re-tuning.

\end{abstract}

\IEEEpeerreviewmaketitle

\section{Introduction} \label{Intro}

In robotics, visual servoing is a well studied research topic \cite{chaumette2006visual,hutchinson1996tutorial} and a well known real-time technique to control the motion of a robot using continuous visual feedback. Such vision based closed loop control increases the accuracy of an overall task, flexibility, functionality and efficiency in robotic automation and safety in collaborative environment while reducing the need for complex fixtures. In conventional visual servoing, frame based cameras are mainly used to extract, track and match visual features by processing images at consecutive frames which causes delays in visual processing and timely robot action. 

\begin{figure}[h!]
 \centering
            \subfloat{\label{fig:sf1}
      \includegraphics[width=0.45\textwidth, height=0.4\textwidth]{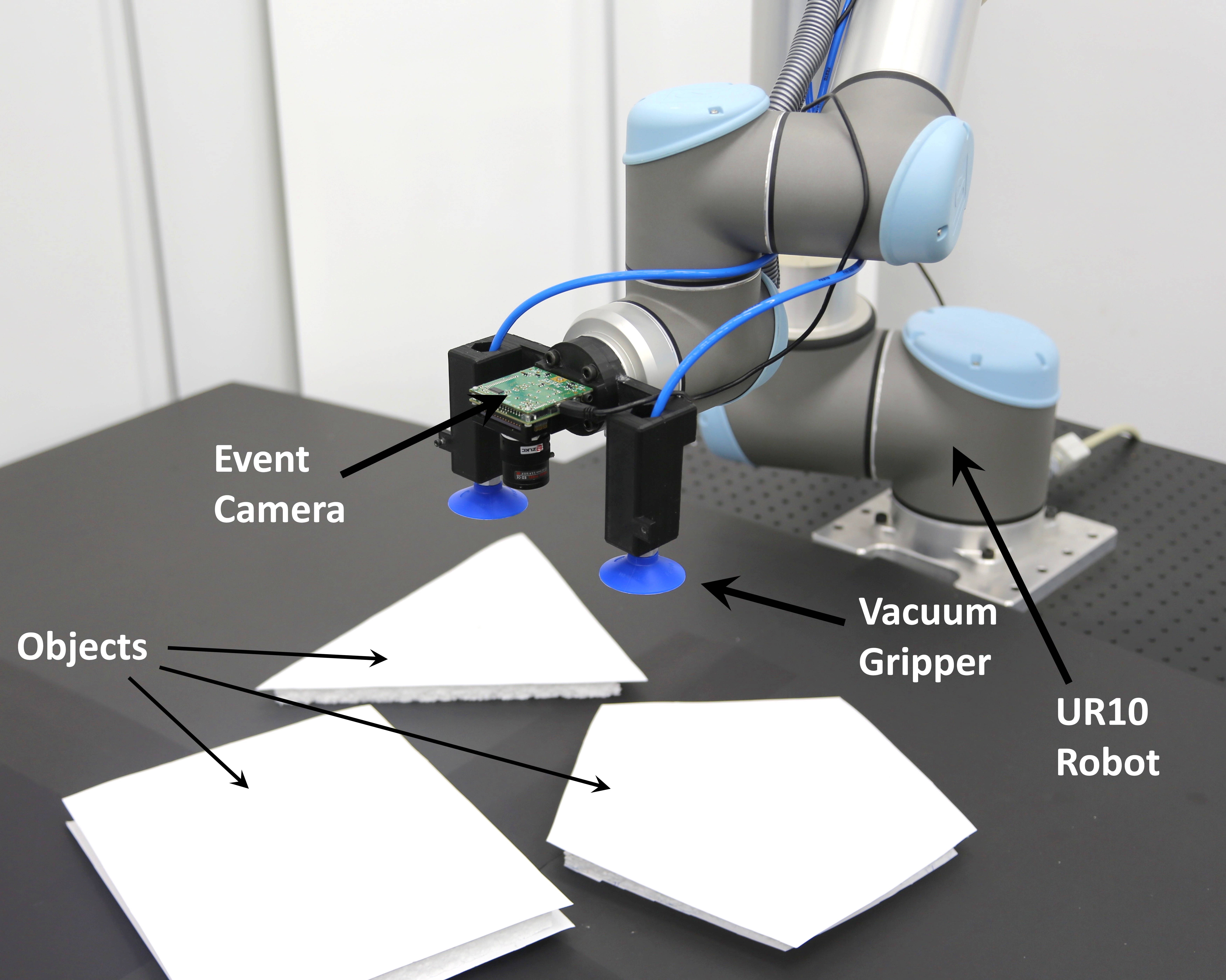}}  \hfill \\
\caption{Event based visual servoing for robotic object manipulation task.}
\label{1_system_real_pic}
\end{figure}

In high-speed applications, the visual information is expected to be fast, efficient, accurate and reliable in providing real-time information of dynamic surroundings. Recently, neuromorphic vision sensors that mimic the neuro-biological architecture of a human retina encodes illumination changes to evolving temporal spikes. Thus, they overcome the limitations of conventional camera and open up a new paradigm shift to visual processing. An event camera attached to the robot end-effector to perform visual servoing is depicted in Fig. \ref{1_system_real_pic}.

Unlike conventional vision sensor which is frame based
and clock driven, neuromorphic vision sensor \cite{vanarse2016review, indiveri2000neuromorphic} is event driven and has low latency, high temporal resolution and wide dynamic range. Moreover, the independent sensor pixels operate asynchronously and in continuous time respond to varying illumination. We exploit this inherent property of the sensor to achieve more efficient and less resource demanding visual servoing to facilitate robotic object manipulation.

In the literature, robotic manipulation pipeline act as a global framework to study such servoing methods \cite{kragic2002survey}. Visual servoing approaches differ by the camera placement, type and number of camera used, 2D or 3D motion command generated, vision algorithm utilized and kinematic and dynamic control strategy deployed. This emphasizes the interdisciplinary efforts for the development of approaches from various fields such as computer vision, control theory, system integration and real-time computation.
Classical approaches are mainly divided into position based visual servoing (PBVS) and image based visual servoing (IBVS). PBVS adopts eye-on-hand configuration and employs the object pose estimated with respect to a calibrated camera as control objective. Thus, they are not able to control the image feature directly, suffer from calibration and estimation errors and requires knowledge of the 3D object model. IBVS on the other hand adopts eye-in-hand configuration and directly use 2D image measures as control objective. They still remain a popular scheme since they exclude the calibration and estimation process. Visual servoing technique has been extensively studied for manipulation applications that accounts rigid, flexible, soft and continuum robot manipulator \cite{cui2020visual,wang2016visual,wang2020eye} .

Similar to the IBVS approach but in the line of event based vision research, we present an event based visual servoing method that adopts the traditional eye-in-hand configuration and process event stream from relative motion to control the motion of the robot. Event camera in such configuration need to act to perceive and perceive to act. We define event based visual servoing as a way to control the motion of the robot using instantaneous spatio-temporal information as feedback. Our approach rely on extraction, robust tracking and matching of event features such as points and lines to reach a desired pose of the event camera, starting from a arbitrary initial pose.

Visual servoing also assist grippers in the grasp alignment process where object shapes and target changes. They even enable a low cost vaccum gripper to align in a range of position and orientation for grasping the object reliably.

\subsection{Contributions} \label{Contributions}
A rich survey on event-based vision is available in \cite{gallego2019event} where several areas relating to robotic applications such as pose tracking, object recognition and tracking, SLAM etc. are reviewed. In the line of event-based vision research, we address the classic problem in robotic grasping and manipulation that is visual servoing. 

In the following the primary contributions of this paper are summarized.

\begin{enumerate}

\item We propose an event based visual servoing (EBVS) method which operates on three layers of active event surface to detect, extract and track high level features and uses a simple control law to dictate the robot motion.


\item We propose a switching strategy within EBVS which enables the robot to explore the work-space to detect key object features and track those features to reach and align the gripper to grasp such that an object manipulation task is facilitated.

\item By constraining the robot with eye-in- hand configuration in a 2D plane, we demonstrate event based visual servoing and gripper alignment to perform a top down grasp using a vaccum gripper which can fit into applications of smart manufacturing.


\end{enumerate}

\section{ Event-based visual Servoing Method} \label{Method}

\begin{figure}[h!]
 \centering
            \subfloat{\label{fig:sf1}
      \includegraphics[width=0.45\textwidth, height=0.25\textwidth]{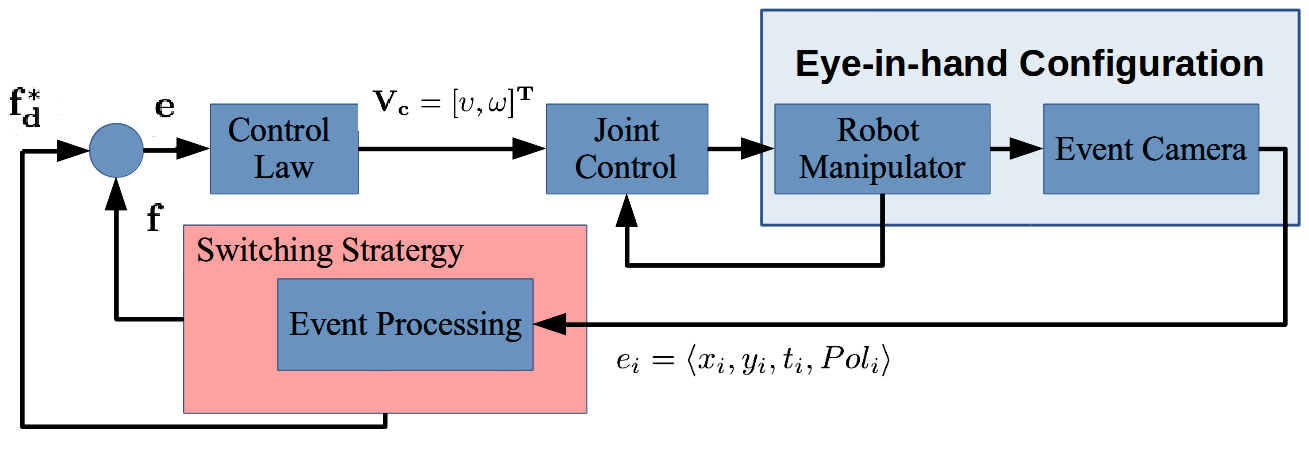}}  \hfill \\
\caption{Block diagram of purely event-based visual control scheme (EVBS)}
\label{2_EBVS_block_diagram}
\end{figure}

An event-based visual control scheme for a robotic manipulator with an eye-in-hand configuration to achieve a manipulation task is illustrated in Fig. \ref{2_EBVS_block_diagram}. Instead of a frame-based camera, an event camera is mounted on the robot's end flange maintaining a relative position with the vaccum gripper. Such setting offer flexibility in viewing the workspace and assistance in grasping. Employing a double loop structure, first, the event stream from neuromorphic vision sensor caused by the dynamic motion is processed to extract high level features.  The switching strategy changes the modes of operation (explore, reach and align) in event-based visual servoing and regulates the feature stream accordingly. Then, these features are used to estimate an error signal between the goal event state and the current state of the feature events. A simple control law ensuring the minimization of the feature error outputs control signal in the form of velocity screw of the event camera. A second loop locally controls and stabilizes the joints of the robotic manipulator. The step by step processing of events, control law and switching strategy is detailed in the following.

\subsection{ Event Processing } \label{Event_processing}
\begin{figure}[h!]
 \centering
            \subfloat[]{\label{fig:sf1}
      \includegraphics[width=0.45\textwidth, height=0.3\textwidth]{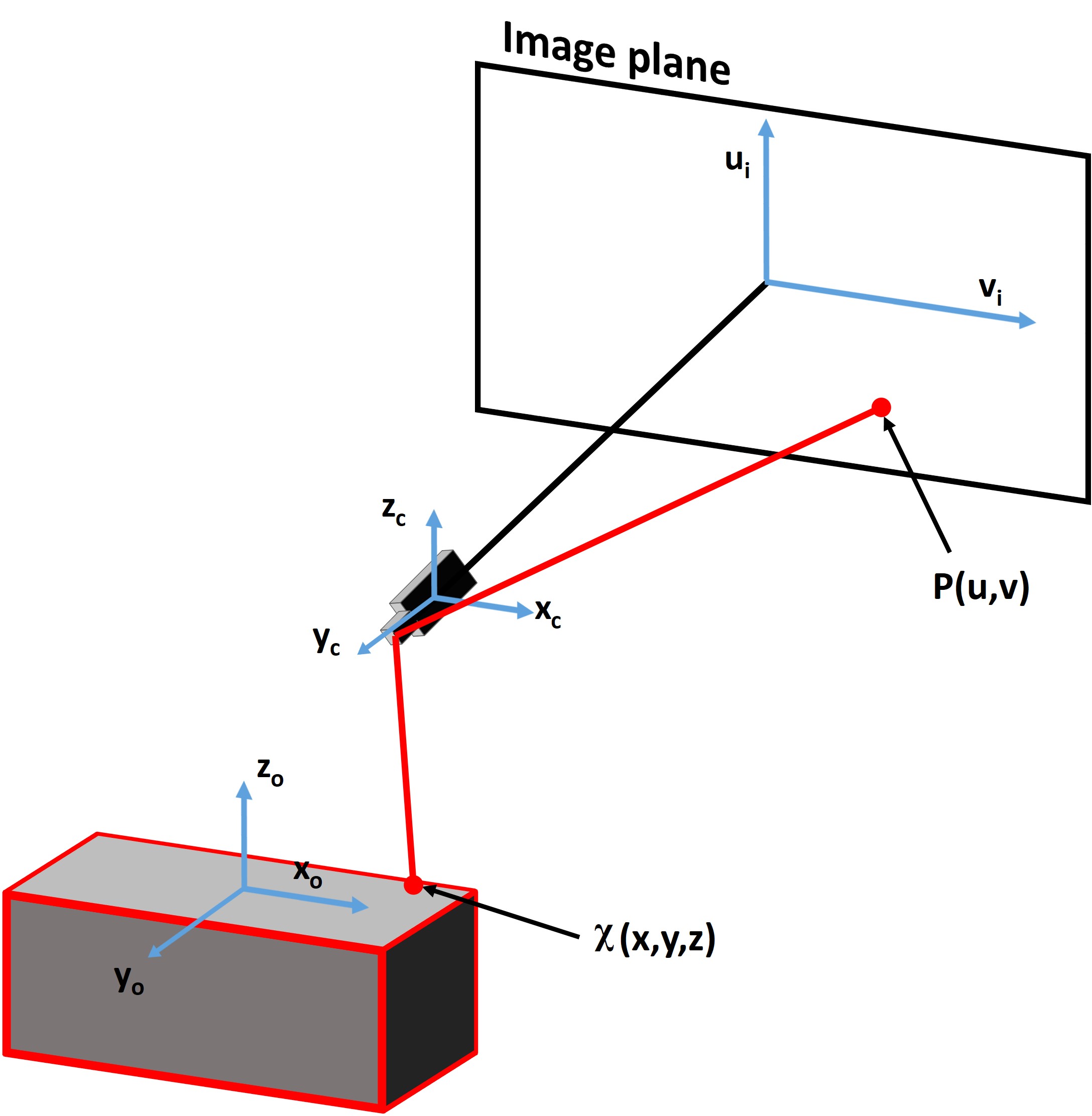}}  \hfill \\
\caption{A moving event camera projects a point illumiation change on a 3D object to a sensor frame.  }
\label{3_3D_to_2D_projection_initial_final_configuration_draw}
\end{figure}

Let us consider a moving event-based camera observing a rigid object placed in a workspace. The movement of the camera generates a stream of events on the sensor plane of the event camera. The standard pin hole model can be still applied in event camera since they use same optics as traditional perspective camera.
The pin hole projection is shown in Fig. \ref{3_3D_to_2D_projection_initial_final_configuration_draw}, mapping a 3D point $\bm{\chi} =[x,y,z]$ into a 2D point $\bm{p} = [u,v]$ on the camera's sensor plane which is expressed in homogeneous coordinates as:
\begin{equation}
\begin{bmatrix}u,v,\bf{1}\end{bmatrix}^T =\quad  \bf{K}   \begin{bmatrix}\bf{I}_{3\times3} \quad \textbf{0}_{3\times1}\end{bmatrix} \begin{bmatrix}
    \bf{R} & \bf{t}  \\
    0 & \bf{1} 
  \end{bmatrix} \begin{bmatrix}f \frac{x}{z}, f \frac{y}{z}, \bf{1}\end{bmatrix}^T
 \label{eq1}
\end{equation}
where $f$ denotes the focal length of the camera, $\bf{K}$ accounts the camera's intrisic components and $\bf{R}$ and $\bf{t}$ refers to  the extrinsic rotational and translation components.

Event cameras represents visual information in terms of time with respect to a spatial reference in the camera-pixel arrays. Pixels in the dynamic vision sensor respond independently and asynchronously to logarithmic brightness changes in the scene. For a relative motion, a stream of events with a microsecond ($\mu s$) temporal resolution and latency is generated, where an event $e =\langle \bm{p},t, Pol \rangle$ is a compactly represented tuple which describes the point $\bm{p}=(u,v)$ in the sensor plane coordinate at time $t$ detailing the brightness increase and decrease by polarity $Pol$. However, analysing a single latest event does not give much information in operational level and exploring all past events is not scalable. 

\begin{figure}[h!]
 \centering
            \subfloat{\label{fig:sf1}
      \includegraphics[width=0.45\textwidth, height=0.3\textwidth]{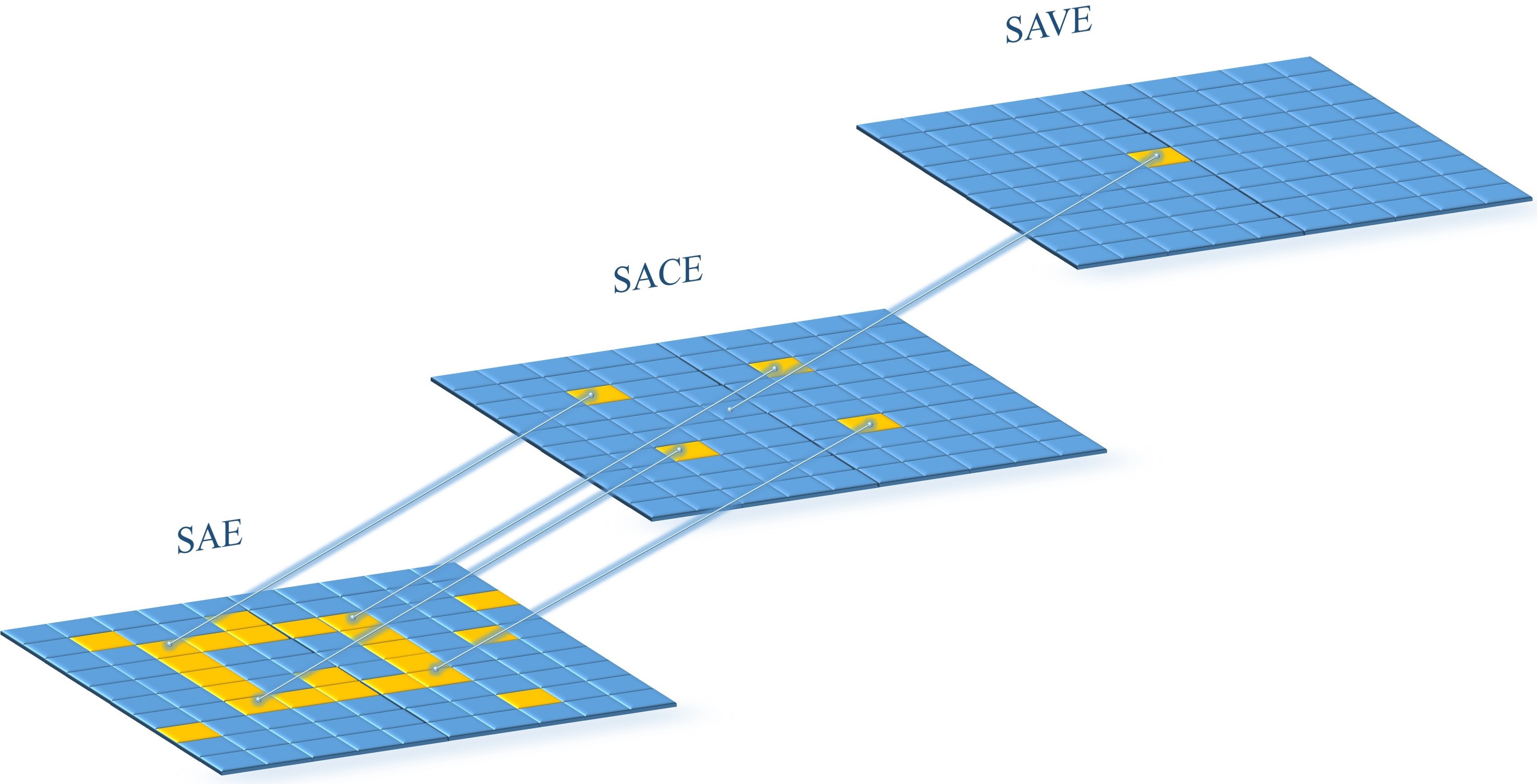}}  \hfill \\
\caption{Three surface layers for active event processing}
\label{4_SAE_3_layers_draw}
\end{figure}

In this work, we consider three sequential layers of surfaces of active events shown in Fig. \ref{4_SAE_3_layers_draw} for performing operations on the evolving temporal data in camera pixel-space to achieve EBVS. The first layer is known as the surface of active events (SAE) where the surface represents the timestamp of a latest event at each pixel from the raw event stream. For each upcoming event, the function $\Sigma_{SAE} : \mathbb{N}^{2} \mapsto \mathbb{R}$ takes the pixel position of a triggered event and assign to its timestamp:

\begin{equation}
\Sigma_{SAE} : (u,v) \mapsto t \: | \: (u,v) \in \mathbb{R} \times \mathbb{R}
 \label{eq1}
\end{equation}

In SAE, we apply feature based algorithms to filter out insignificant events and extract highly informative events such as corners. The second layer is the surface of active corner events (SACE) which maps the pixel position of recent corner events to its time stamp, where we extract the center of the object by robustly localizing the corner events. The object center is the extracted high level feature that is a virtual event and not an actual event used in visual servoing. Moreover, we introduce random and goal state events and consider them as virtual events. The third layer is the surface of active virtual events (SAVE) that maps the extracted and artificially induced virtual events pixel position to its timestamp, where the contiguity of the high level feature is analysed for switching the control objectives. EVBS modes of operations such as exploration, reaching and grasping are determined by the SAVE.

\subsection{ Feature Detection } \label{Feature_detection}


\begin{figure*}[h!]
 \centering
      \subfloat[]{\label{fig:sf1}
      \includegraphics[width=0.3\textwidth, height=0.2\textwidth]{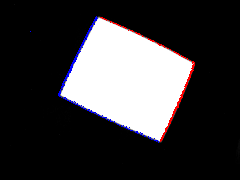}}  \qquad
      \subfloat[]{\label{fig:sf1}
      \includegraphics[width=0.3\textwidth, height=0.2\textwidth]{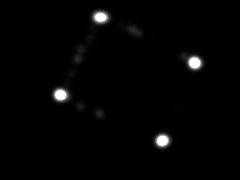}}  \qquad
      \subfloat[]{\label{fig:sf1}
      \includegraphics[width=0.3\textwidth, height=0.2\textwidth]{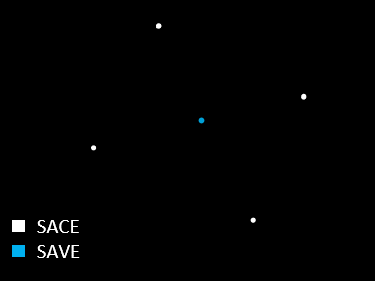}}  \qquad
\caption{Event Based Feature Detection: (a) Rendered event frame (30ms) (b) Weighted Corners heatmap (c) Features events extracted in SACE and SAVE.}
\label{5_detection_heatmap_SACE_draw}
\end{figure*}

In conventional image processing, Harris detector is one of the most widely used technique that detects features such as corner, edge and flat points based on Strong intensity variation in a local neighborhood. This feature detector is known for its efficiency, simplicity and in-variance to scaling, rotation and illumination.  Unlike conventional camera that records large amount of redundant data in sequence of frames, the DVS records only the changes in the visual scene as stream of events characterized by the pixel positions and its timestamps and does not include intensity measures. Therefore the frame based harris detector cannot be directly applied on the SAE. Event-based adaptation of harris detector is proposed in \cite{vasco2016fast} and \cite{mueggler2017fast} where each upcoming event is directly processed. Their method binarizes the SAE by the newest $N$ events for the whole image plane or locally around the current event.
Let $\Sigma_{b}$ be a binary surface locally centered around the latest event where 1 and 0 indicates the presence and absence of an event. The gradient is computed on the binary surface with $5 \times 5$ sobel operator as

\begin{equation}
I_{x}=\Sigma_{b} G_{x} ; I_{y}=\Sigma_{b} G_{y}
 \label{eq1}
\end{equation}
 Compute Harris matrix
\begin{equation}
H_{m}= \sum_{e\in \Sigma_{b}} Gu(e) \nabla I_{e} \nabla I_{e}^{T}
 \label{eq1}
\end{equation}
 Compute Harris score
\begin{equation}
H_{s}= det{(H_{m})} - k.trace (H_{m})^{2}
 \label{eq1}
\end{equation}

The Harris feature detector mainly relies on the analysis of the eigenvalues of the auto-correlation matrix. If the Harris score is large positive value, the event is classified as corner whereas a negative value is considered as edge. The rest of the events which are in-between is considered as flat points. In our case, the adapted e-Harris detector \cite{mueggler2017fast} is used to detect corner events from locally perceived information that is  independent of the scene and sensor size. Selected corner threshold of $HC_{th} = 5$, buffer of latest events $N = 20$ and a patch of $9 \times 9$ pixels gave the best performance over a wide variety of data-sets.

Whenever a corner event $p_{hc}: (x_{hc}, y_{hc})$ is detected it is projected in the SACE. To cluster these events into object corners and minimize the influence of noise events, consecutive corner points are concatenated to form a heat-map of corner locations. A heat-map matrix \(H \in \mathbb{R} \times \mathbb{R}\) is introduced for this purpose. Whenever a new corner event is received, the elements of \(H\) are updated as:

\begin{equation}
H_{i, j}(t^+) = H_{i, j}(t^-) + \alpha  e^{-0.5  \times ((x_{hc} - i)^{2} +(y_{hc} - j)^{2}) / \sigma^{2} } 
\label{eq_heatmap}
\end{equation}

Where $x_c$ and $y_c$ represents the coordinates of an recent corner event, \(\alpha\) is a scaling factor and \(\sigma\) is the standard deviation of the incoming corner event which dictates the area of effect each event has on the heat-map. 

To keep only the recent events relevant to the process of detecting the object corners, the heat-map is continuously updated with time as indicated by eq. \ref{eq_heatmap_time}, where \(\tau\) is a time constant dictating the period of influence for each corner point and \(t_c\) is the timestamp of the last received corner event.

\begin{equation}
H(t^{+}) = e^{-\tau (t_c - t)} \times H(t^{-})
\label{eq_heatmap_time}
\end{equation}

As such, the heat-map \(H\) represents spatio-temporal patterns in the corner events. The corners of the object are then obtained from these patterns by detecting the local peaks if these peaks exceed a minimum threshold of 0.7. Local maxima are obtained by dilating the heat-map with a window size of $10 \times 10$ and extracting locations where the original heat-map is equal to the dilated heat-map. Fig. \ref{5_detection_heatmap_SACE_draw} shows an example corner heat-map along with its local peaks for a sample object place in the camera's field of view. Let $S=\{(x_{c}^{0},y_{c}^{0}),...,(x_{c}^{n},y_{c}^{n}) \}$ be the set of local peaks from the heat map of $n$ corner points. We consider centroid of the object as the high level feature projected in SAVE for tracking operations in visual servoing which is computed as

\begin{equation}
(x_{v},y_{v}) =(\frac{1}{n} \sum_{i=0}^{n} x_{c}^{i}, \frac{1}{n} \sum_{i=0}^{n} y_{c}^{i})
 \label{eq1}
\end{equation}

\begin{figure*}[h!]
 \centering
            \subfloat{\label{fig:sf1}
      \includegraphics[width=\textwidth, height=0.2\textwidth]{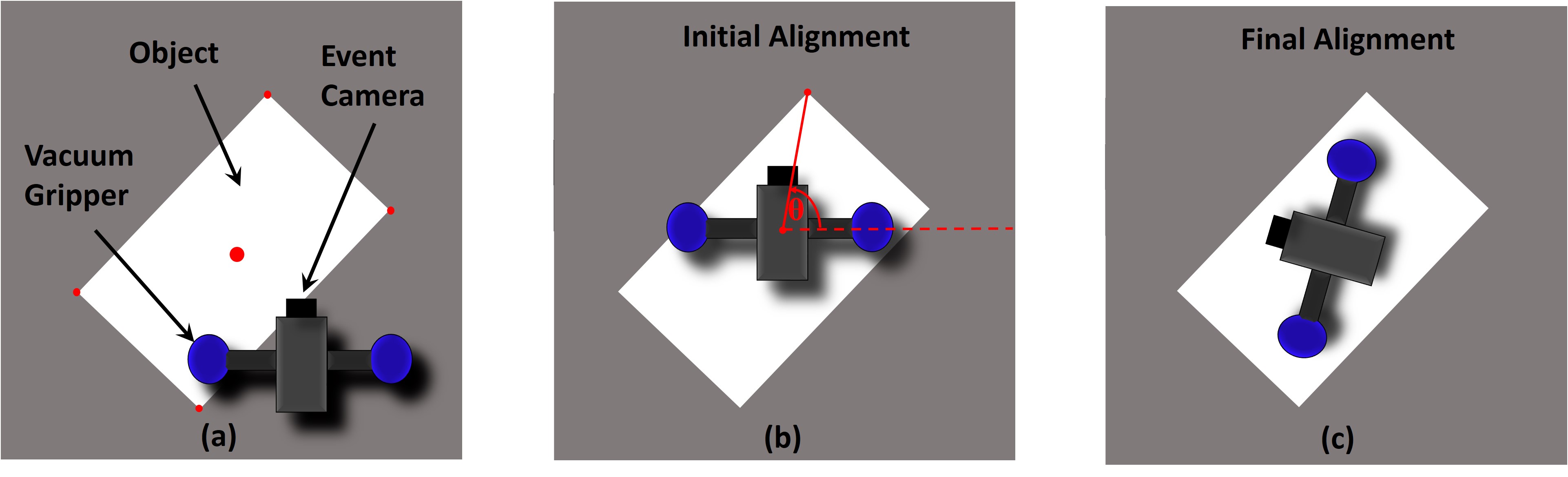}}  \hfill \\
\caption{Constraint set for gripper alignment after servoing.}
\label{7_alignment_draw}
\end{figure*}

\begin{figure}[h!]
 \centering
            \subfloat{\label{fig:sf1}
      \includegraphics[width=0.45\textwidth, height=0.25\textwidth]{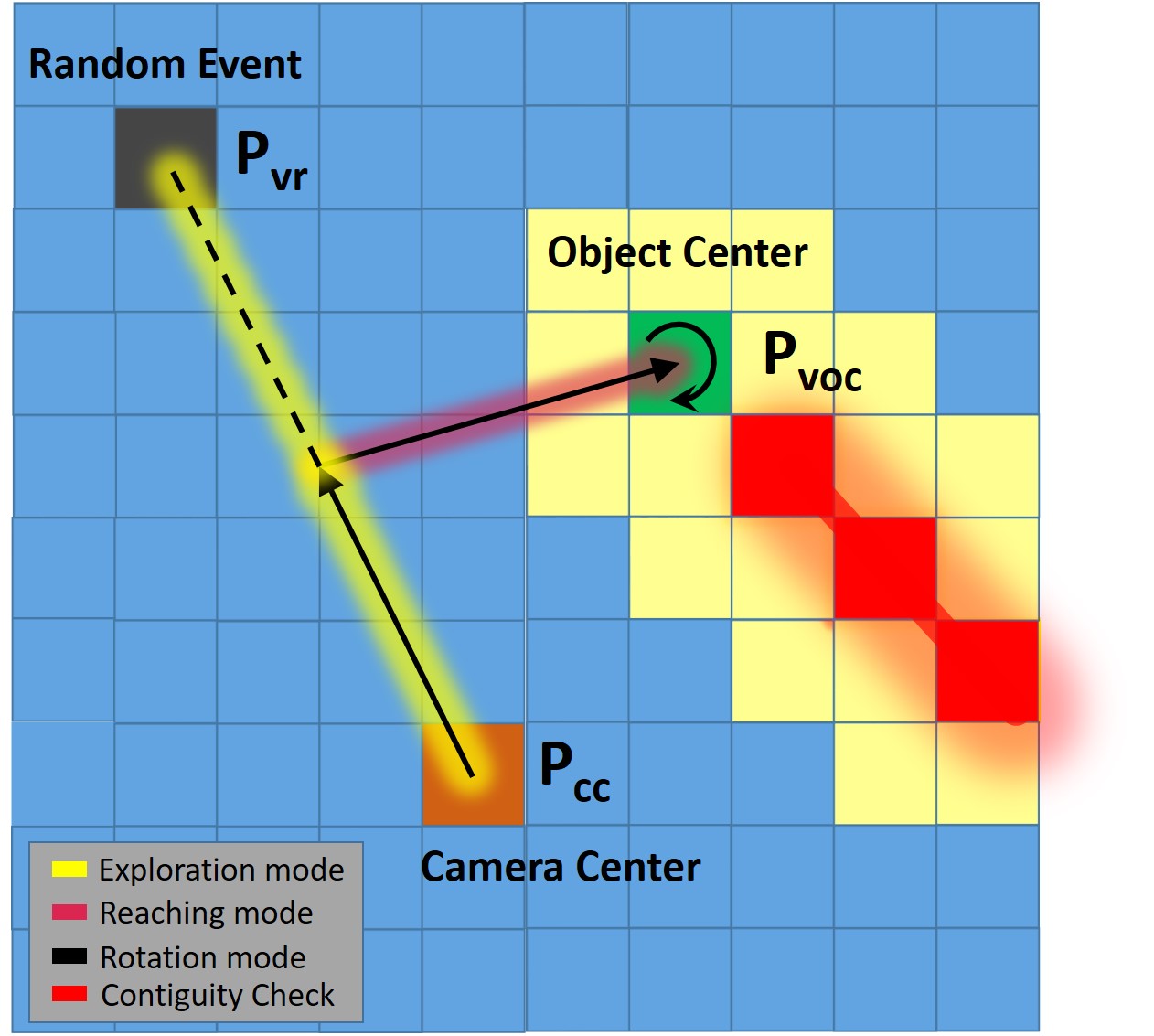}}  \hfill \\
\caption{Illustration of switching stratergy that explore to detect, track to reach and align to grasp.}
\label{6_stratergy_SAVE_draw}
\end{figure}

\subsection{ Feature tracking } \label{Visual_servoing}

Let $\bf{f_{d}^{*}}$ denote the desired feature events triggered in SAVE (for example the center of the sensor plane) and $\bf{f}$ gives the coordinates of the detected high level feature events such as object center, both expressed in pixel units. The linear $\bf{\upsilon}$ and angular $\bf{\omega}$ velocity of the camera is represented as $\bf{V_{c}}= (\upsilon,\omega)$. The primary goal in EVBS is to compute the camera velocity $\bf{V_{c}}$ such that the error $\bf{e}= \bf{f}-\bf{f_{d}^{*}}$ is minimized. The relationship between the velocity of the feature events and the camera velocity is given by

\begin{equation}
\dot{\bf{f}} =\bf{L} \bf{V_{c}}
 \label{eq1}
\end{equation}
in which $\bf{L} \in \mathbb{R}^{k \times 6}$ is the feature Jacobin. Moore-penrose pseudo-inverse $ \bf{L}^{\dagger}= (\bf{L}^{T}\bf{L})^{-1}\bf{L}^{T}$ is used when it is full rank. To control 6 DOF, atleast three feature points are necessary, $\bf{L}$ can be stacked together in a composite form so to achieve.  $\bf{V_{c}}$ is the input to the robot controller ensuring an exponential decrease of the feature error ($\dot{\bf{e}} = \lambda \bf{e}$) and the control law is expressed as

\begin{equation}
\bf{V_{c}} = - \lambda \bf{L}^{\dagger} \bf{e}
 \label{eq1}
\end{equation}

As the end-effector moves towards the object, the location of the object's corners and centroid in the sensor plane must be updated. For this purpose, a simple moving average approach is adopted. For every new $p_{hc}$ detected by the e-Harris algorithm, the closest object corner $p^i_{c}:(x_c^i, y_c^i) \in S $ is determined. $p^i_{c}$ is then updated as:

\begin{equation}
p^i_{c}(t^+) = 0.9 \times p^i_{c}(t^-) + 0.1 \times p_{hc}
 \label{eq_MM_tracking}
\end{equation}

Whenever the SACE is updated, the SAVE is also updated accordingly, leading to a refined estimate of the object's centroid.

Due to its simplicity, tracking corners using the moving average approach is much faster than the heat-map corner detection; making it more suitable for high speed application. However, it is prone to errors if tracking of one corner is lost. To account for such cases, corner tracking is regularly checked against heat-map corner detection at an interval of 0.3s, if considerable discrepancies were found over multiple timesteps, the system reverts into heat-map-based corner detection mode. 







\subsection{ Gripper Alignment to Grasp } \label{Visual_servoing}
Once the robot tracks and reaches the object's center, the orientation of the grippers is adjusted to achieve a stable grasp. A target orientation \(\theta\) is defined such that the two gripping points are aligned with a virtual line connecting the object centroid \(p_v: (x_v, y_v)\) in the SAVE with a corner point \(p^i_{c}\) in the SACE. To maximize the stability of the grip, \(p^{i}_{c}\) is selected as the corner point furthest from the centeroid. \(\theta\) is hence computed as:

\begin{equation}
\theta = atan2(y^{i}_{c} - y_{v},  x^{i}_{c} - x_{v})
 \label{eq_orientation}
\end{equation}

Fig. \ref{7_alignment_draw} shows the alignment process where the grippers are rotated at a constant angular velocity until \(\theta\) is within an admissible range.

\subsection{ Switching Strategy } \label{Visual_servoing}
The switching strategy enables the robot to explore, reach and grasp in the process of event based visual servoing. In Fig. \ref{6_stratergy_SAVE_draw}, the switching operation is illustrated in the surface of active virtual events. Let $p_{cc}$ be the artificially induced desired event representing the central pixel of the camera at the starting position, $p_{vr}$ a random feature event and $p_{voc}$ is the extracted recent feature event representing the center of the object. First, a virtual event $p_{vr}$ is triggered to motivate the robot to explore the workspace and detect object feature $p_{voc}$. The highlighted yellow color indicates the pathway chosen by the robot in the exploration phase. While tracking, the contiguity of $p_{voc}$ is analysed. Once the count of contiguous pixel crosses above a threshold (e.g. 3). The robot changes its coarse of action and tracks $p_{voc}$ to minimize the error. The highlighted pink color indicates the new pathway to reach the object center. Switching can happen even in the reaching phase due to detection issues and contiguity breakdown. However, the strategy gives the robot the capability to recover and reach the desired feature. Finally, the robot aligns the gripper to perform a stable grasp. The switching function can be expressed as

\begin{equation}
\bf{f}(\bf{P}(t)) = =
    \begin{cases}
      \bold{P} = \bold{p}_{vr}^{i} & ,\text{if contiguity in $\bold{p}_{voc}^{j,..,n} \ | \ j<3$}\\
      \bold{P} = \bold{p}_{voc}^{j} & ,\text{if contiguity in $\bold{p}_{voc}^{j,..,n}\ | \ j>3$}\\
      \bold{P} = \bold{p}_{va}^{k} & ,\text{$\bold{p}_{voc}^{j} = \bold{P}_{cc} $}
    \end{cases}  
 \label{eq1}
\end{equation}

\begin{figure}[!th]
 \removelatexerror
\begin{algorithm}[H]
\DontPrintSemicolon

  \KwInput{Stream of events $e_{i} = \langle x_{i},y_{i},t_{i},Pol_{i}\rangle$, e-Harris threshold: $HC_{th}$ , Contiguity threshold $C_{th}$}
  \KwOutput{Camera velocity $\bf{V_{c}}= (\upsilon,\omega)$.}
\KwConfig{Eye-in-hand}
Initialize three layers of surface of active events SAE, SACE and SAVE.\\
Initialize a desired feature event (eg: center of the sensor plane) in SAVE\\
Initialize switching strategy \\

    \For{ each $e_{i}$}{ 

Detect corner features in SAE by applying e-Harris with $HC_{th}$.\\
Extract weighted corners in SACE using heat-maps.\\
Compute object centroid from weighted corners events in SACE.\\
Monitor and operate in SAVE.\\

\If {Contiguity count $<$ $C_{th}$}
{
Initialize a random desired event in the SAVE.\\
Engage visual servoing to the random feature event.\\
Detect and track object feature events in SAVE.\\

}

\If {Contiguity count $>$ $C_{th}$}
{
Track the object centroid events. \\

}

\If {camera center $=$ object center}
  {
Align gripper orientation for a stable grasp.\\
Move to the pre-grasp pose and execute grasp.\\
   	}

   	}

\caption{Event-based Visual Servoing}
\end{algorithm}
\end{figure}

%
%

\begin{figure*}
\setlength{\tempwidth}{.3\linewidth}
\settoheight{\tempheight}{\includegraphics[width=\tempwidth]{example-image-a}}%
\centering
\hspace{\baselineskip}
\columnname{Side View}\hfil
\columnname{Top View}\hfil
\columnname{Heat Map}\\
\rowname{Exploration Phase}
\subfloat[Ready to explore]{\includegraphics[width=0.3\textwidth, height=0.21\textwidth]{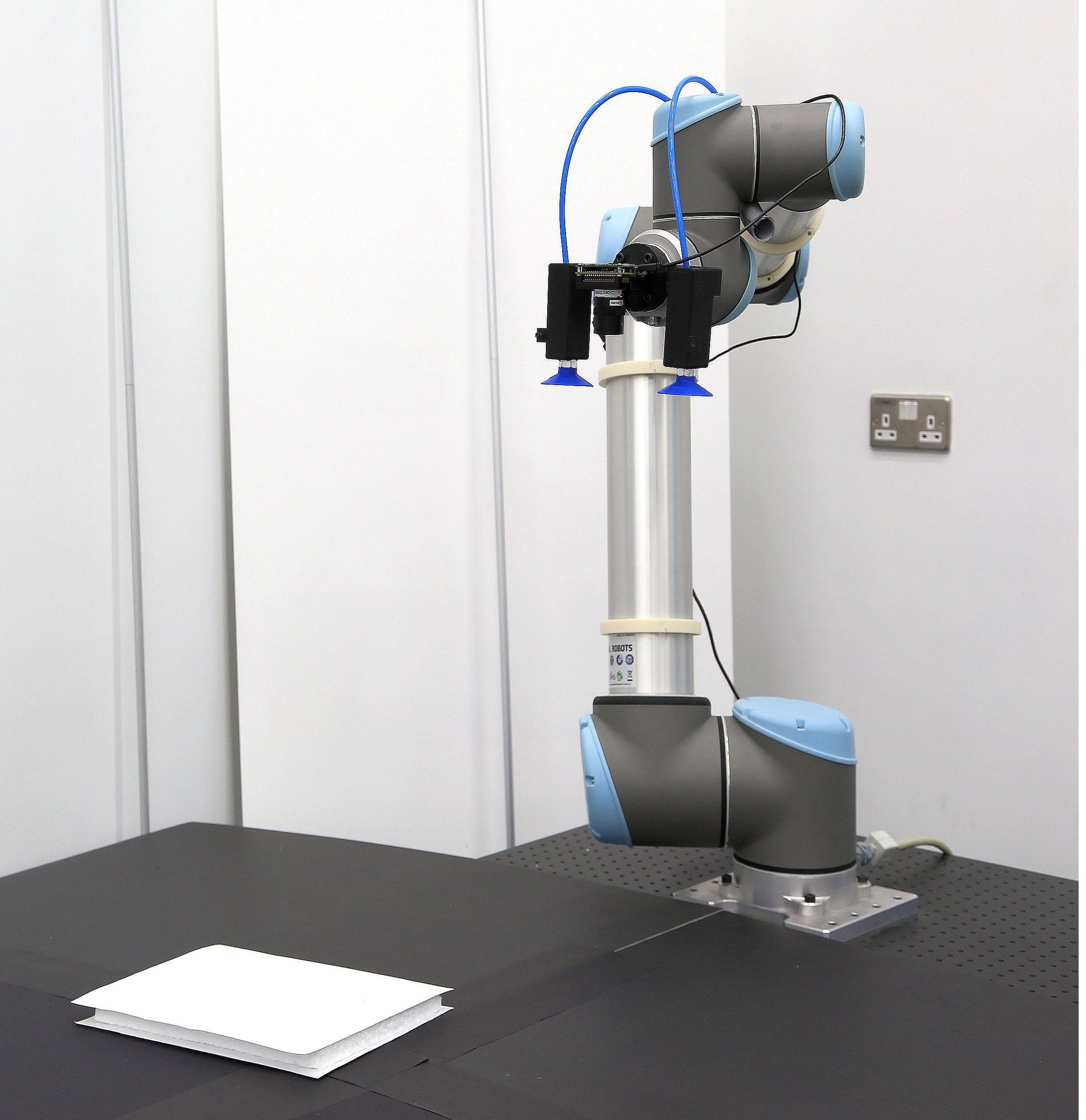}}\label{a}\hfil
\subfloat[Random virtual event ($p_{vr}$) generated]{\includegraphics[width=0.3\textwidth, height=0.21\textwidth]{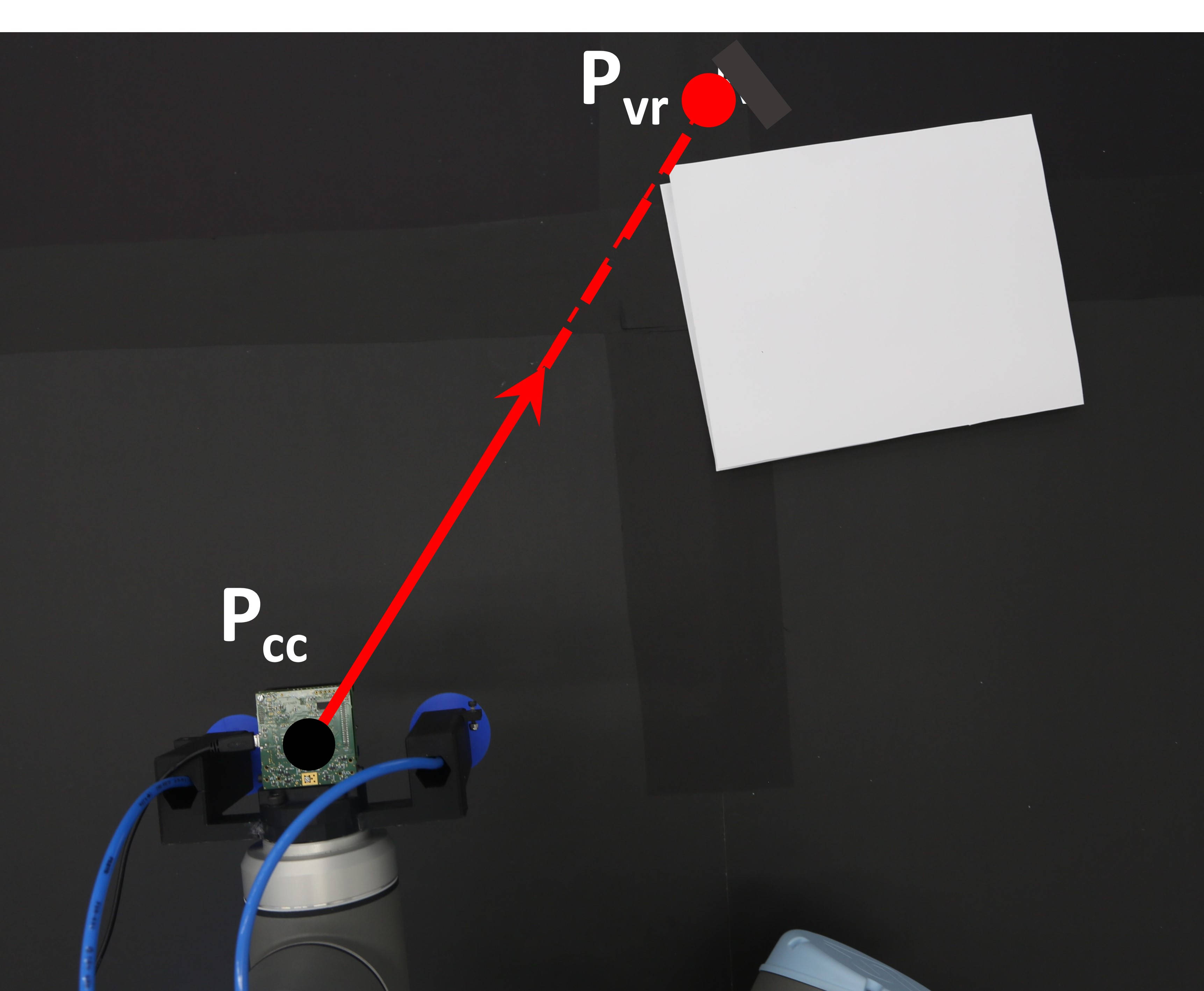}}\label{b}\hfil
\subfloat[Object corners(white) and $p_{vr}$ (red) depicted in heat map]{\includegraphics[width=0.3\textwidth, height=0.22\textwidth]{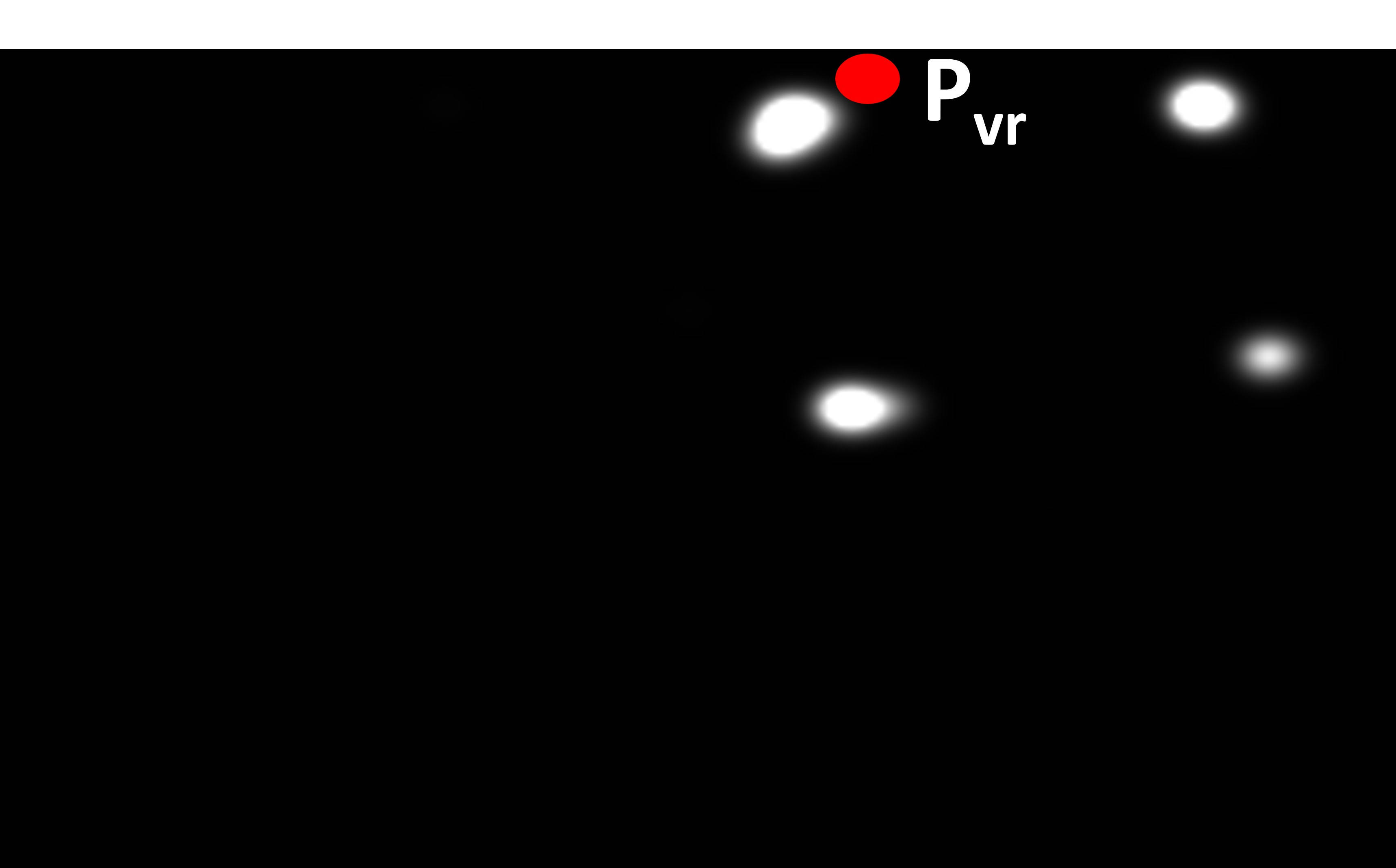}}\label{c}\\
\rowname{Reaching Phase}
\subfloat[Change coarse of action]{\includegraphics[width=0.3\textwidth, height=0.21\textwidth]{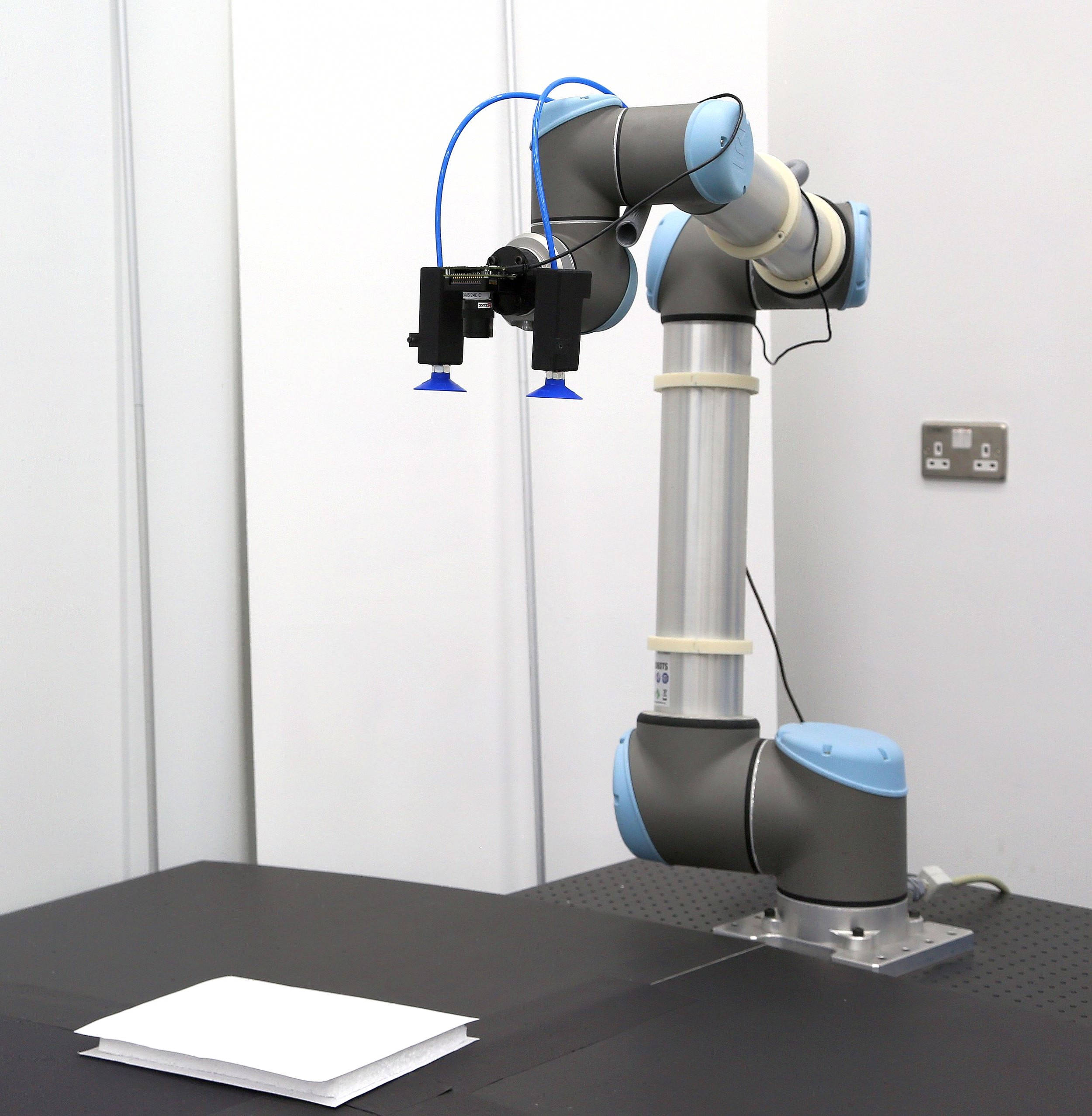}}\label{d}\hfil
\subfloat[Object center ($p_{voc}$) detected after contiguity check ]{\includegraphics[width=0.3\textwidth, height=0.21\textwidth]{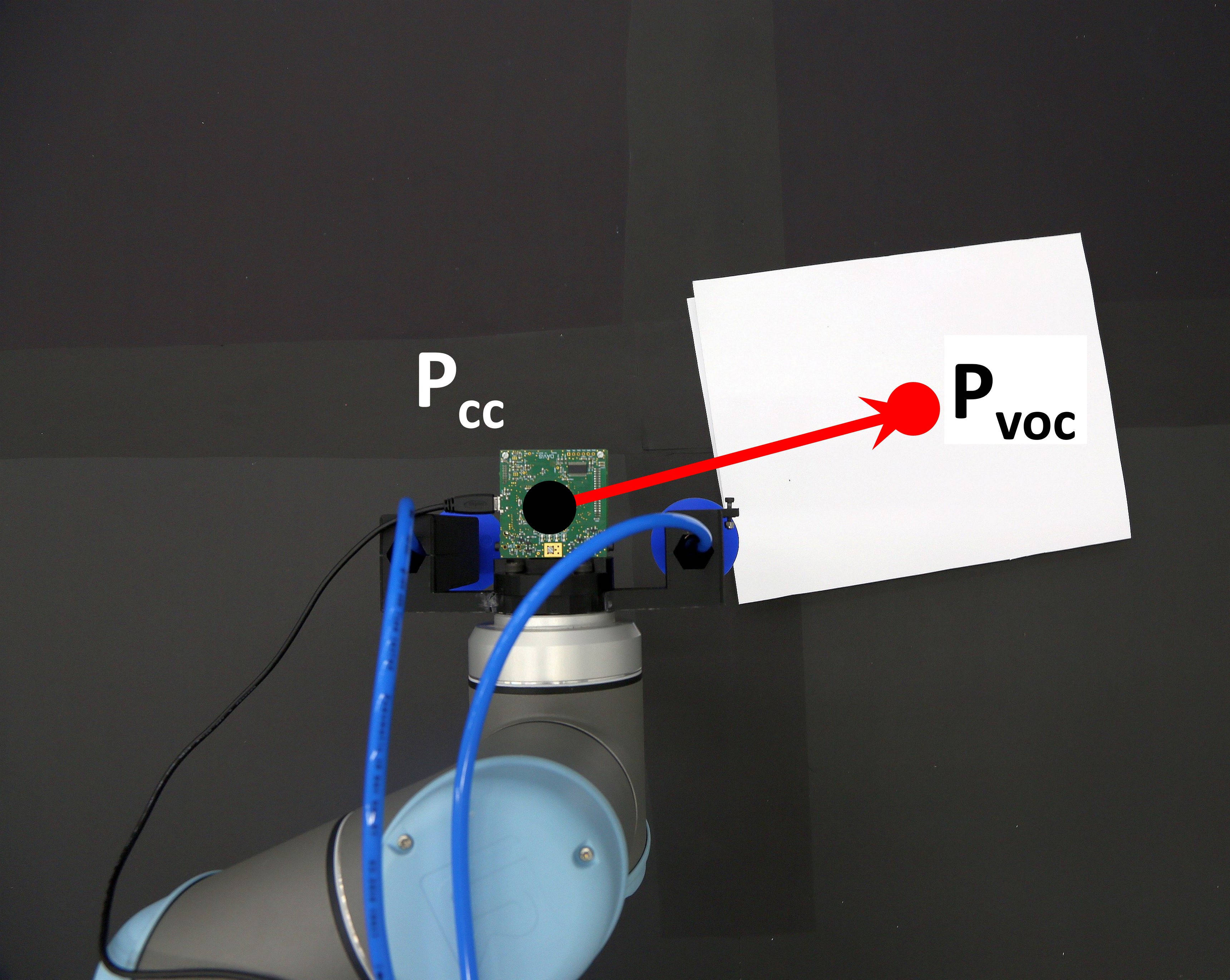}}\label{e}\hfil
\subfloat[Object corners(white) and $p_{voc}$ (red) depicted in heat map]{\includegraphics[width=0.3\textwidth, height=0.21\textwidth]{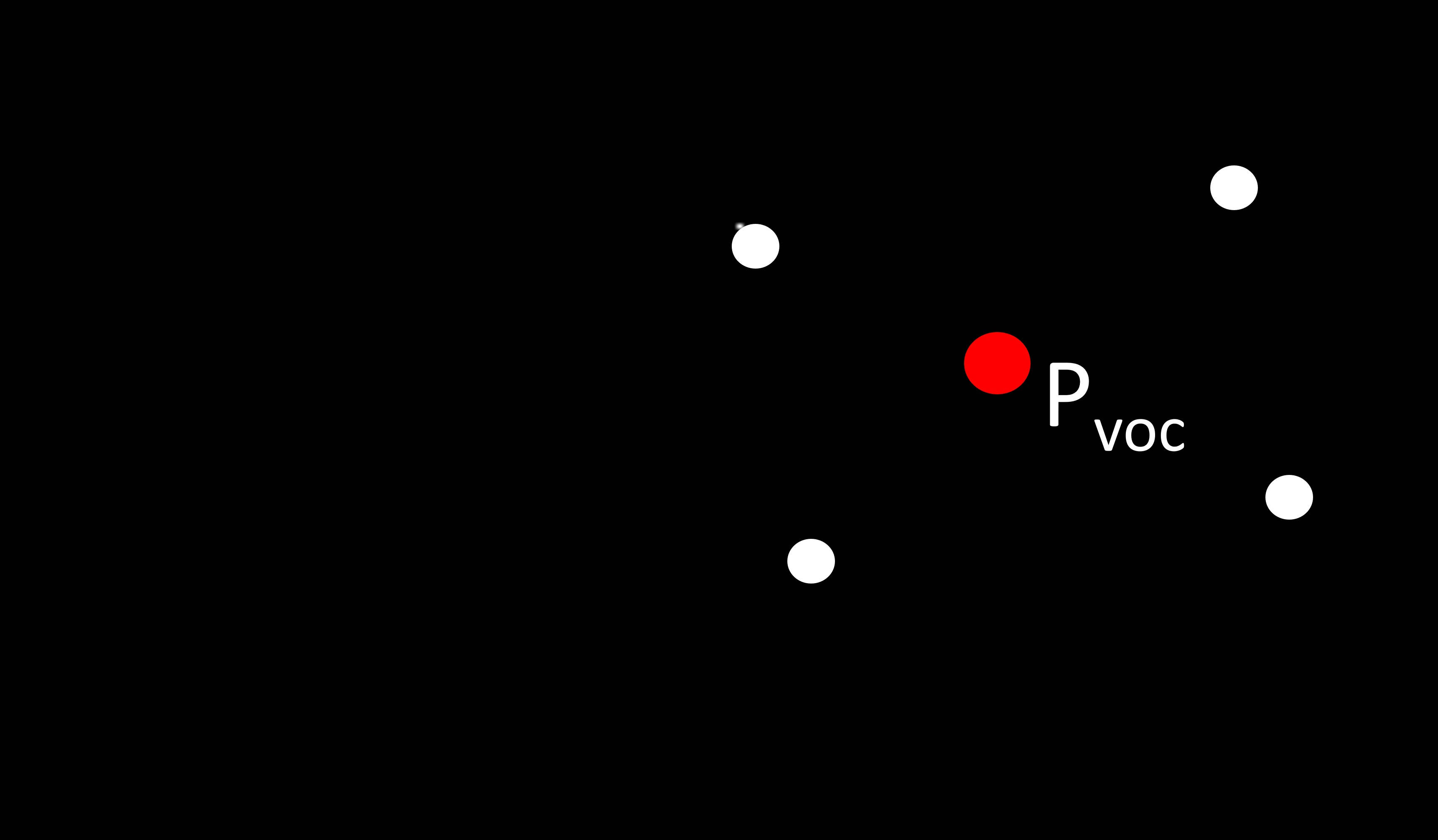}}\label{f}\\
\rowname{Alignment Phase}
\subfloat[Initial alignment]{\includegraphics[width=0.3\textwidth, height=0.21\textwidth]{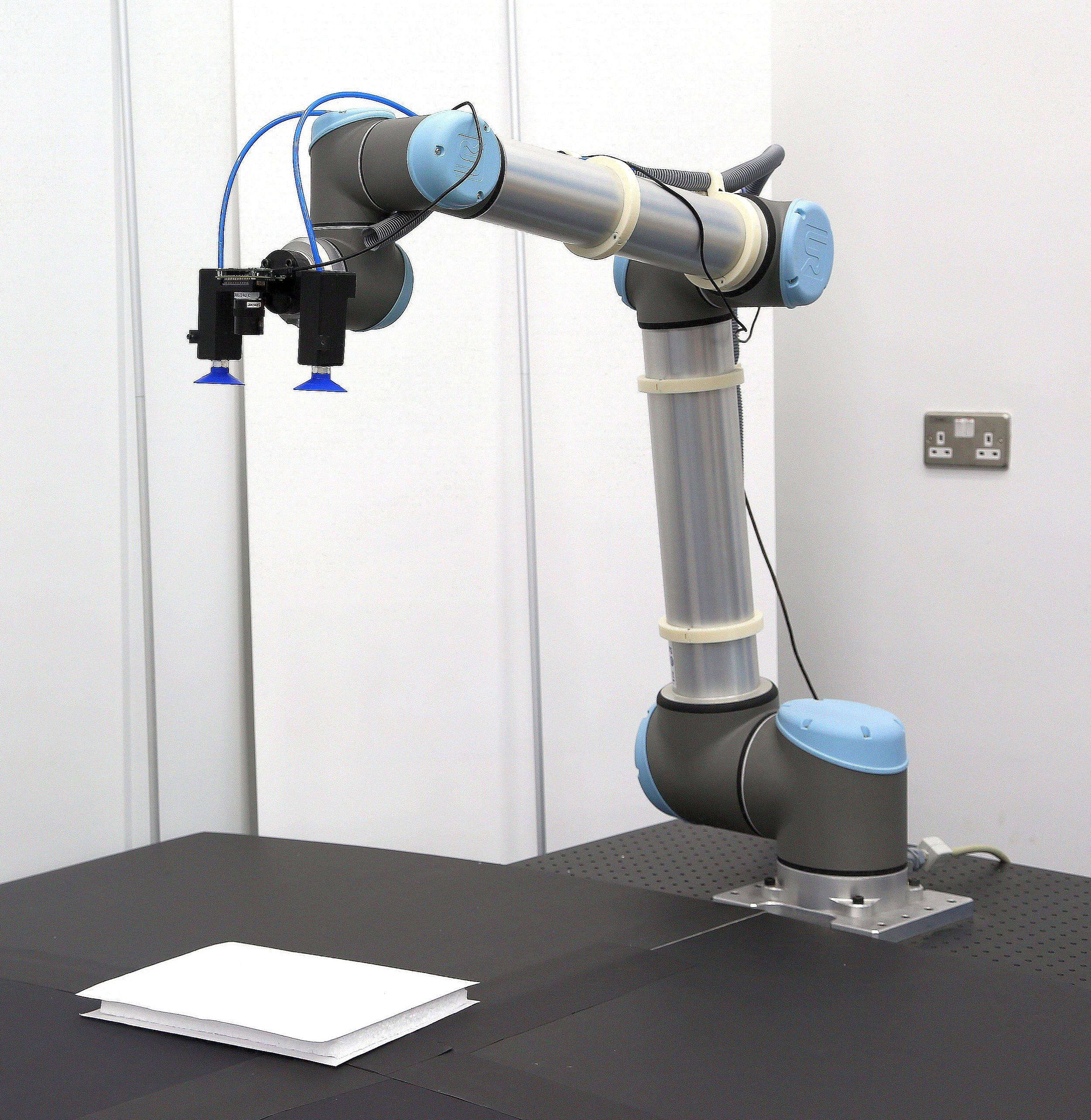}}\label{g}\hfil
\subfloat[Camera and object center matched]{\includegraphics[width=0.3\textwidth, height=0.21\textwidth]{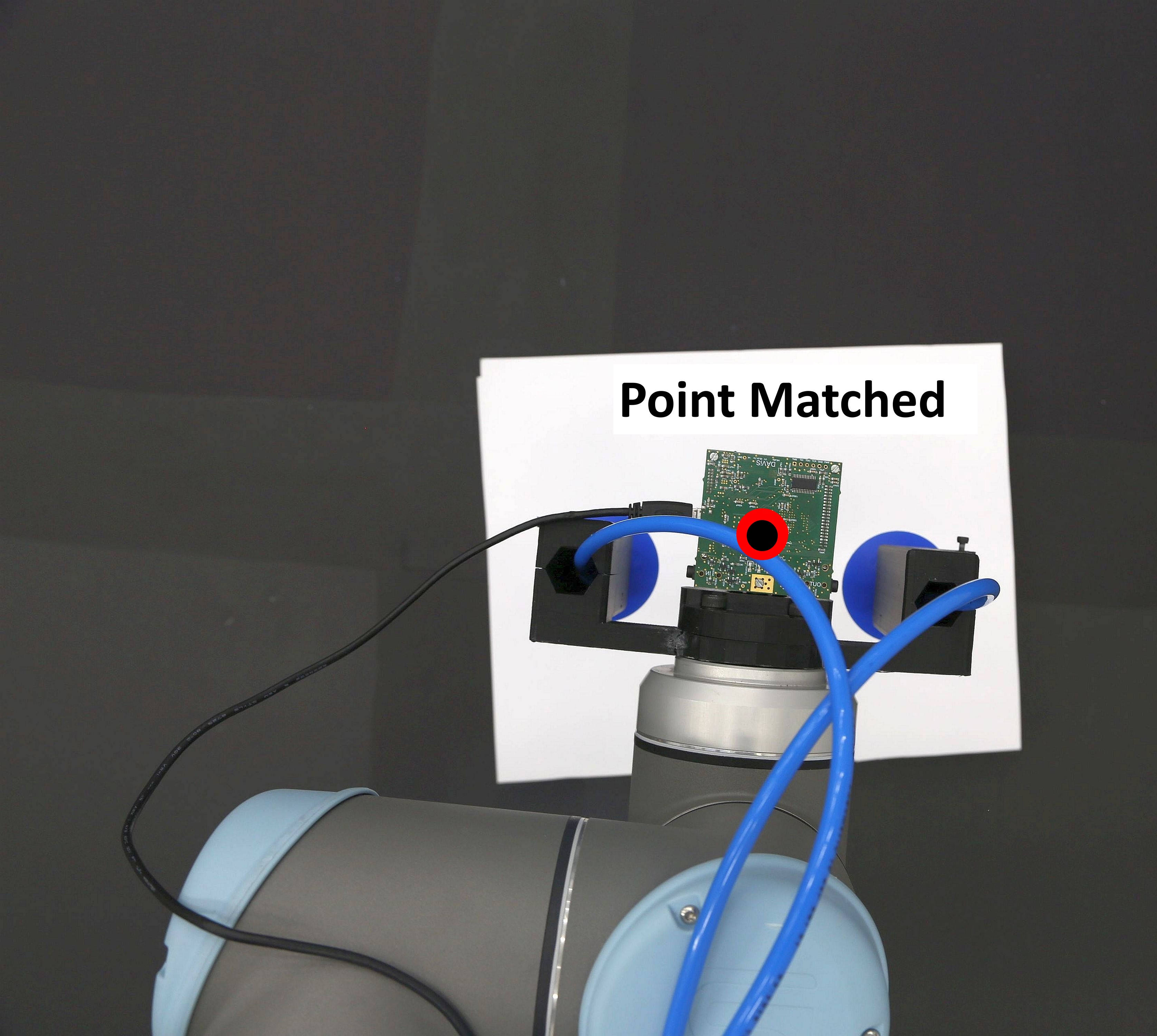}}\label{h}\hfil
\subfloat[Object corners(white) and matched featues (red)]{\includegraphics[width=0.3\textwidth, height=0.21\textwidth]{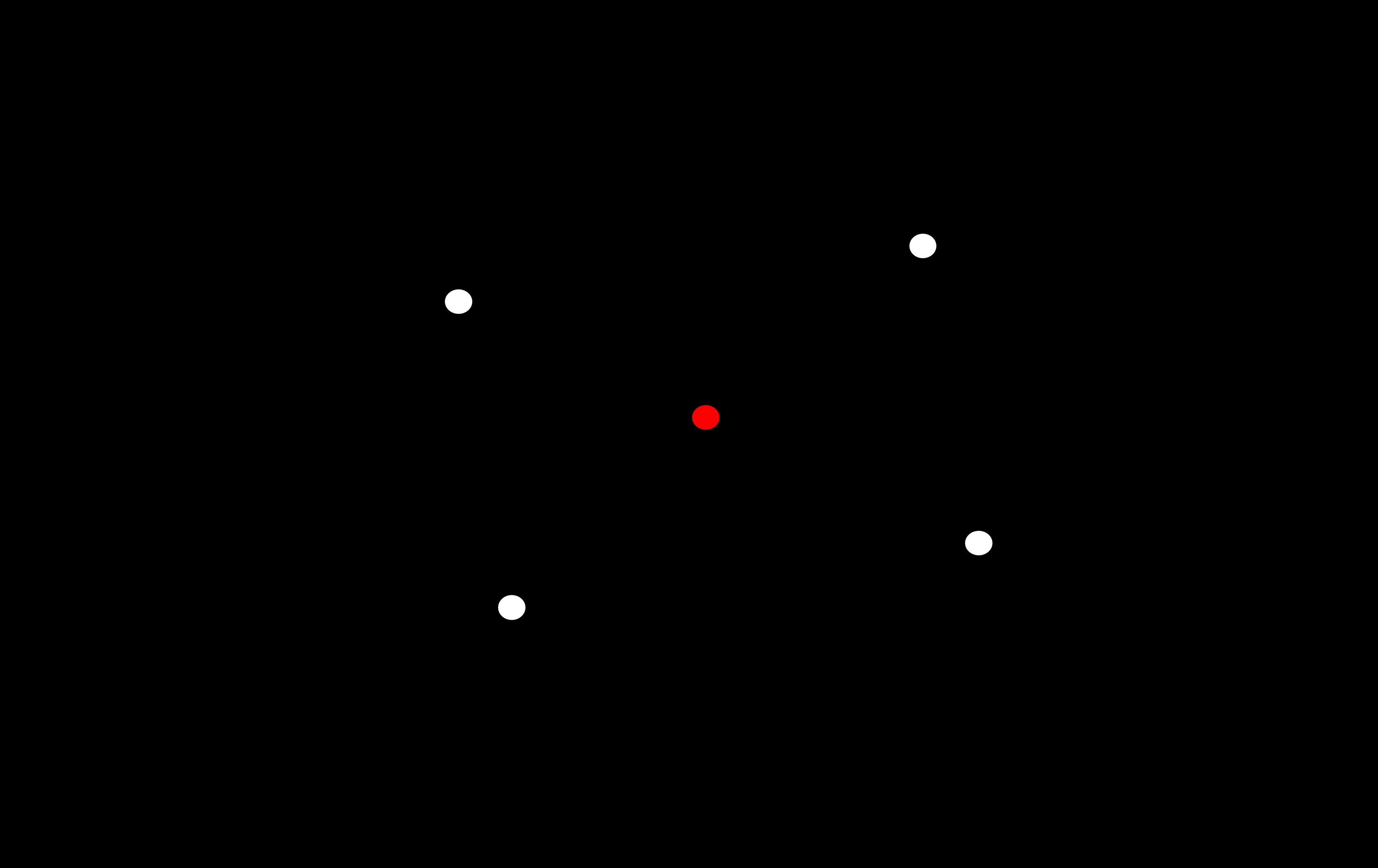}}\label{i}\\
\rowname{Grasping Phase}
\subfloat[Final alignment]{\includegraphics[width=0.3\textwidth, height=0.21\textwidth]{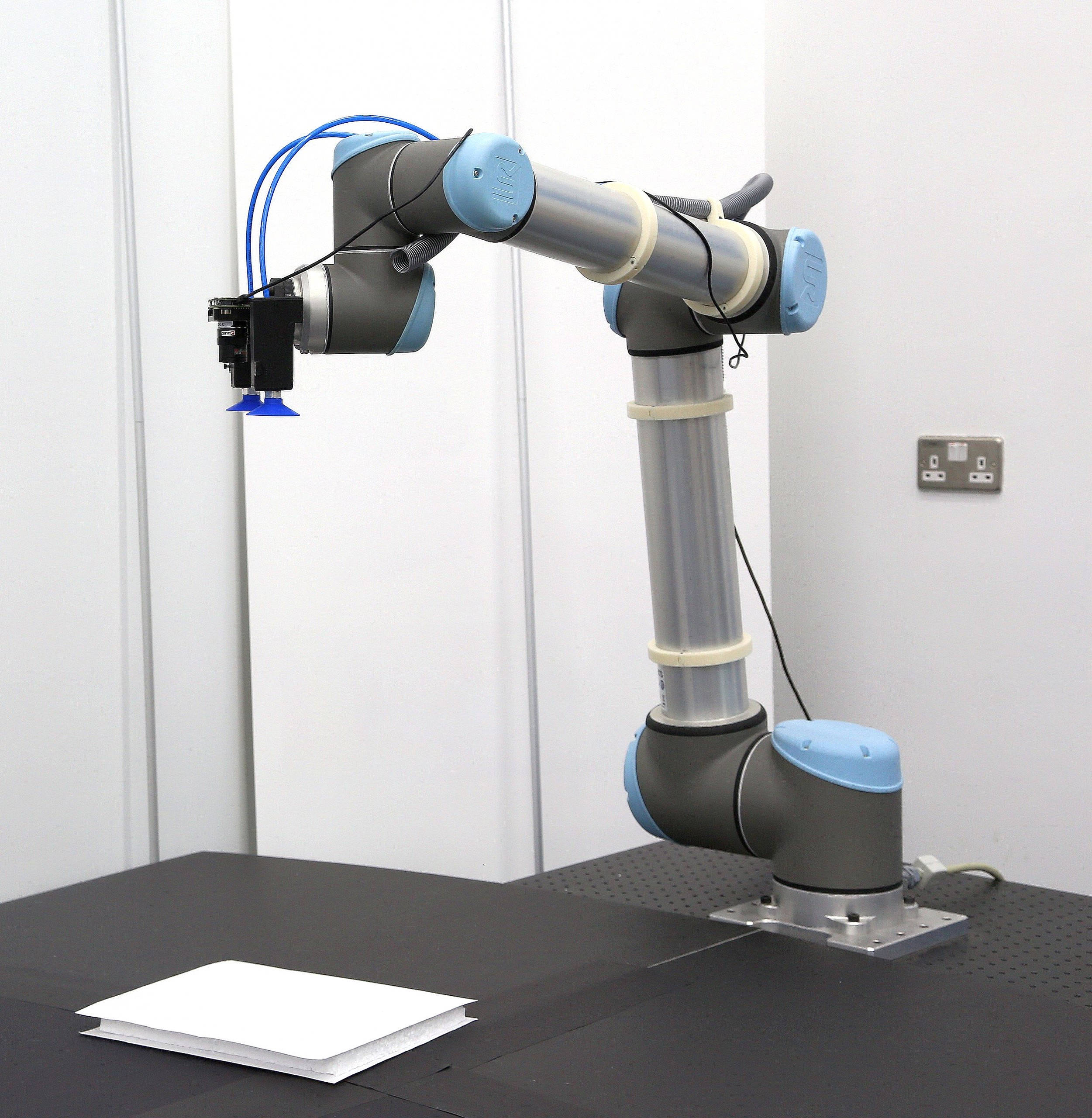}}\label{j}\hfil
\subfloat[Gripped based alignment for stable grasp]{\includegraphics[width=0.3\textwidth, height=0.21\textwidth]{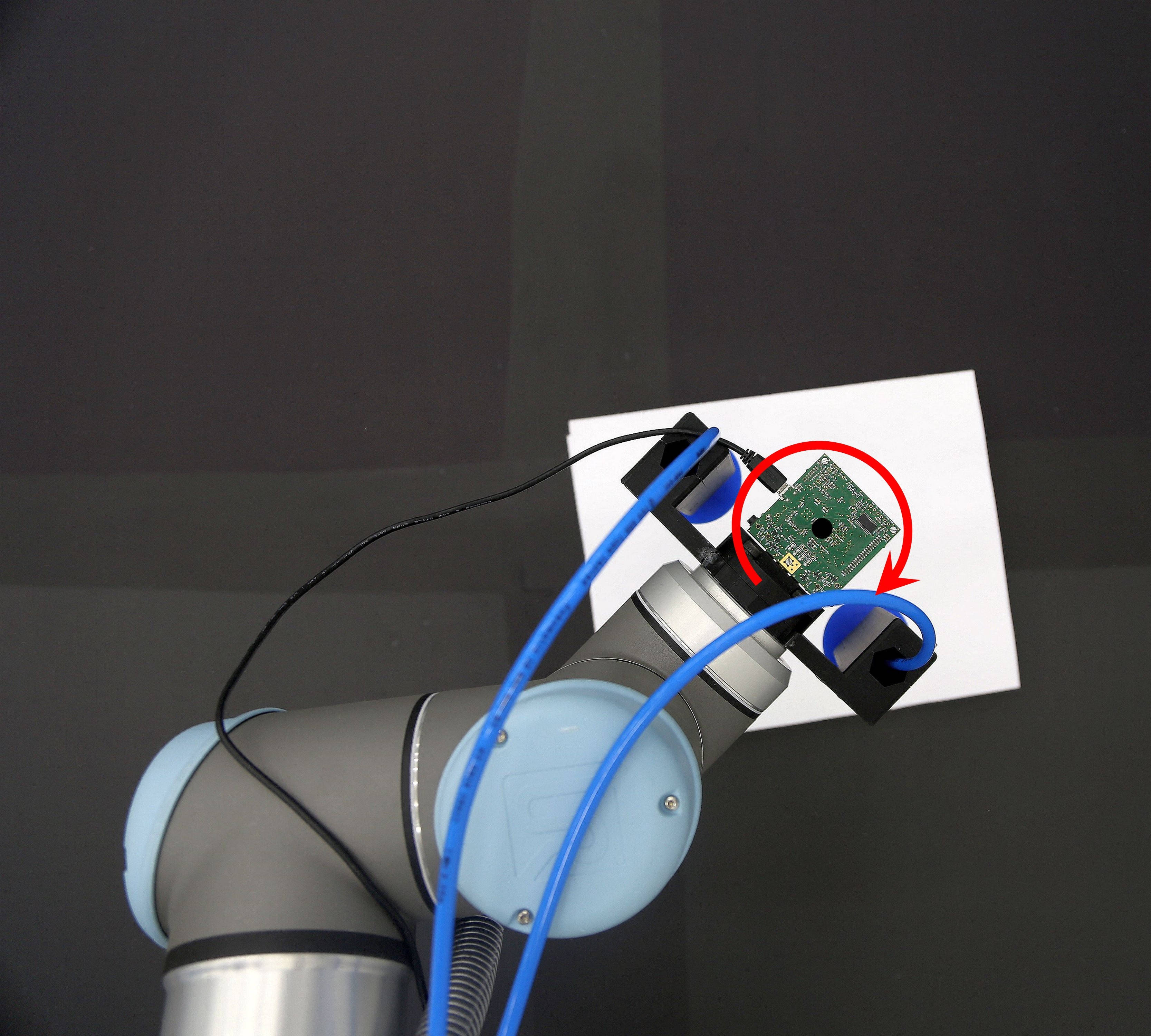}}\label{k}\hfil
\subfloat[Farthest corner point used in alignment]{\includegraphics[width=0.3\textwidth, height=0.21\textwidth]{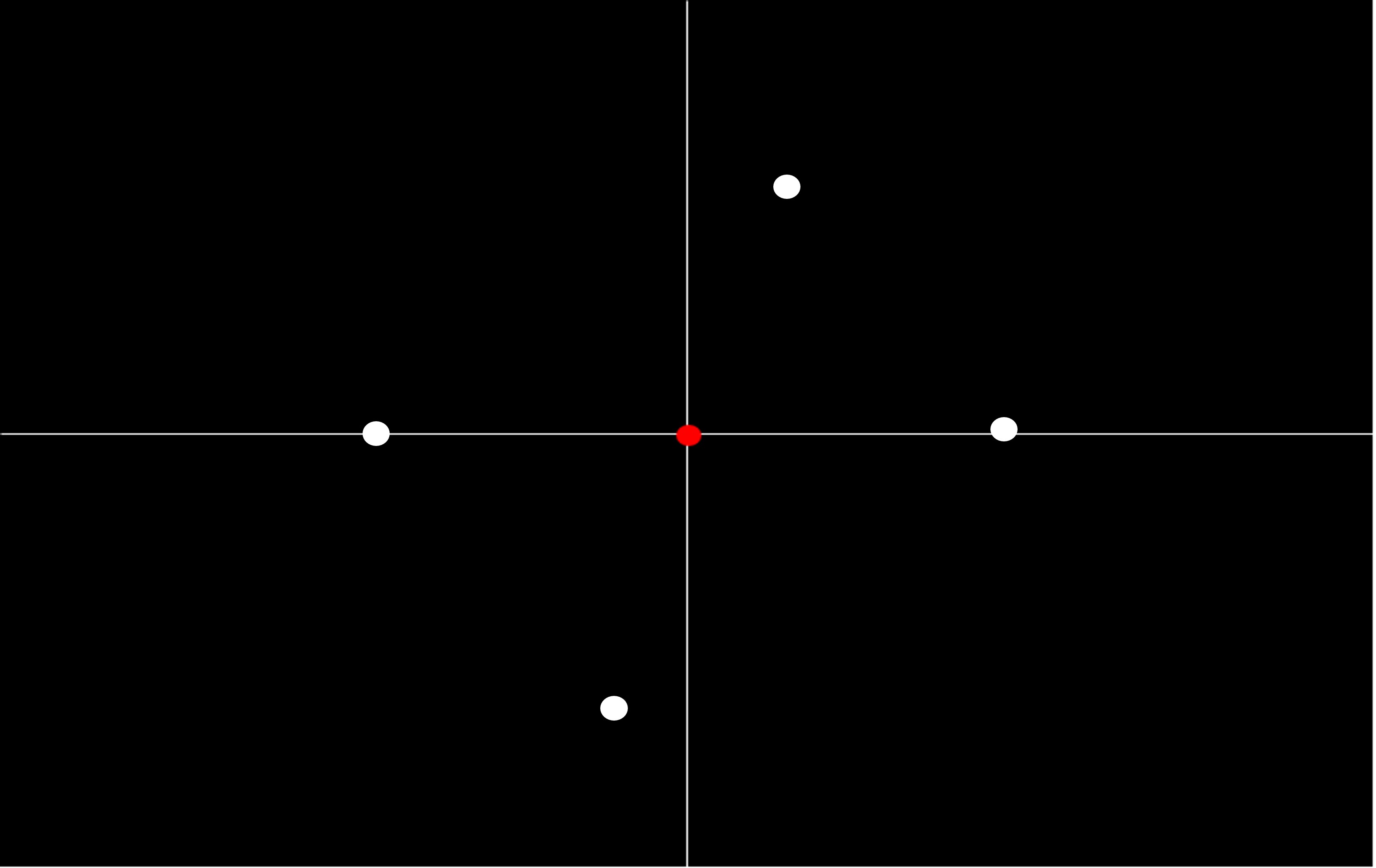}}\label{l}\\
\rowname{Manipulation Phase}
\subfloat[Pick]{\includegraphics[width=0.3\textwidth, height=0.21\textwidth]{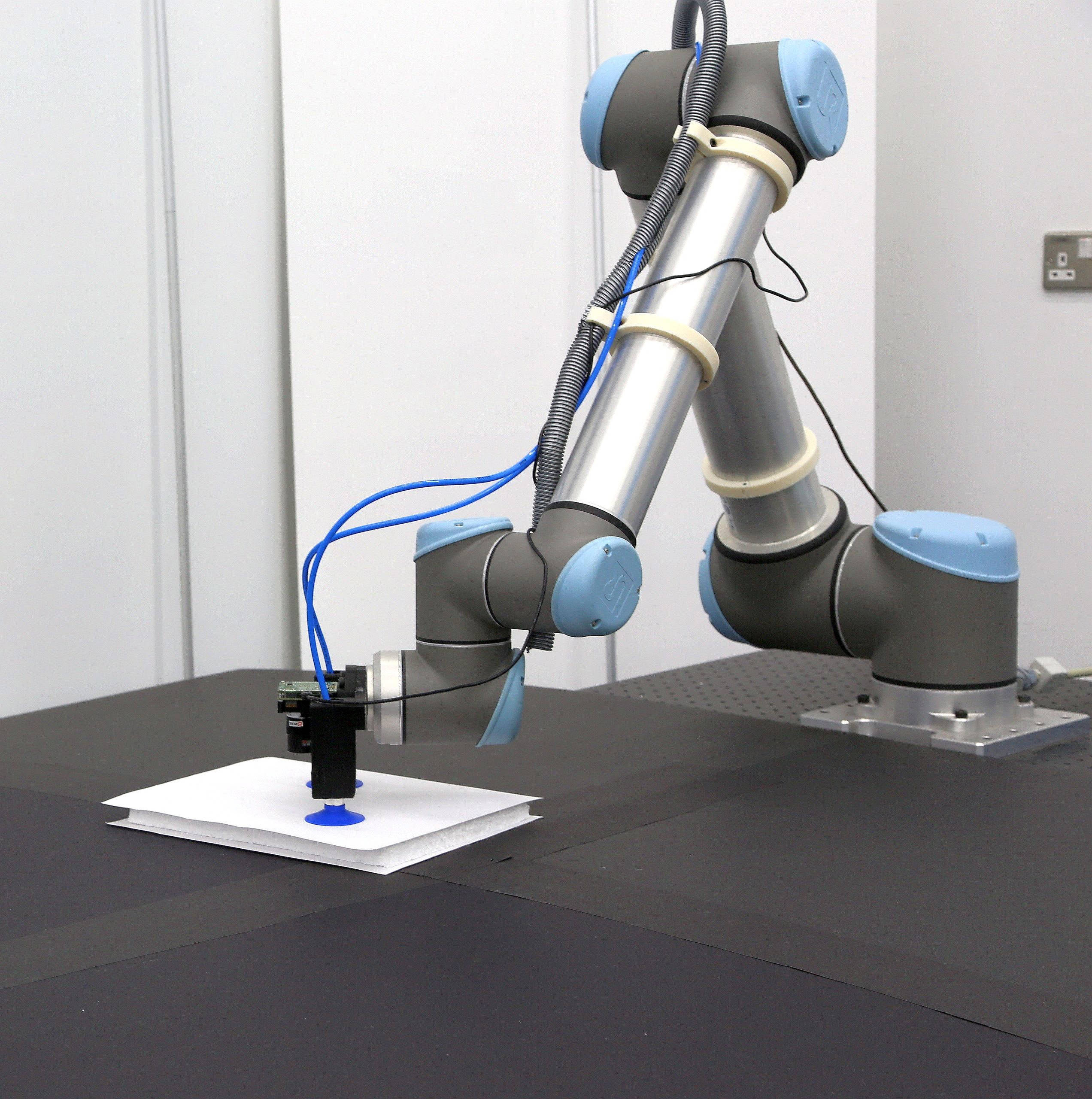}}\label{m}\hfil
\subfloat[Lift]{\includegraphics[width=0.3\textwidth, height=0.21\textwidth]{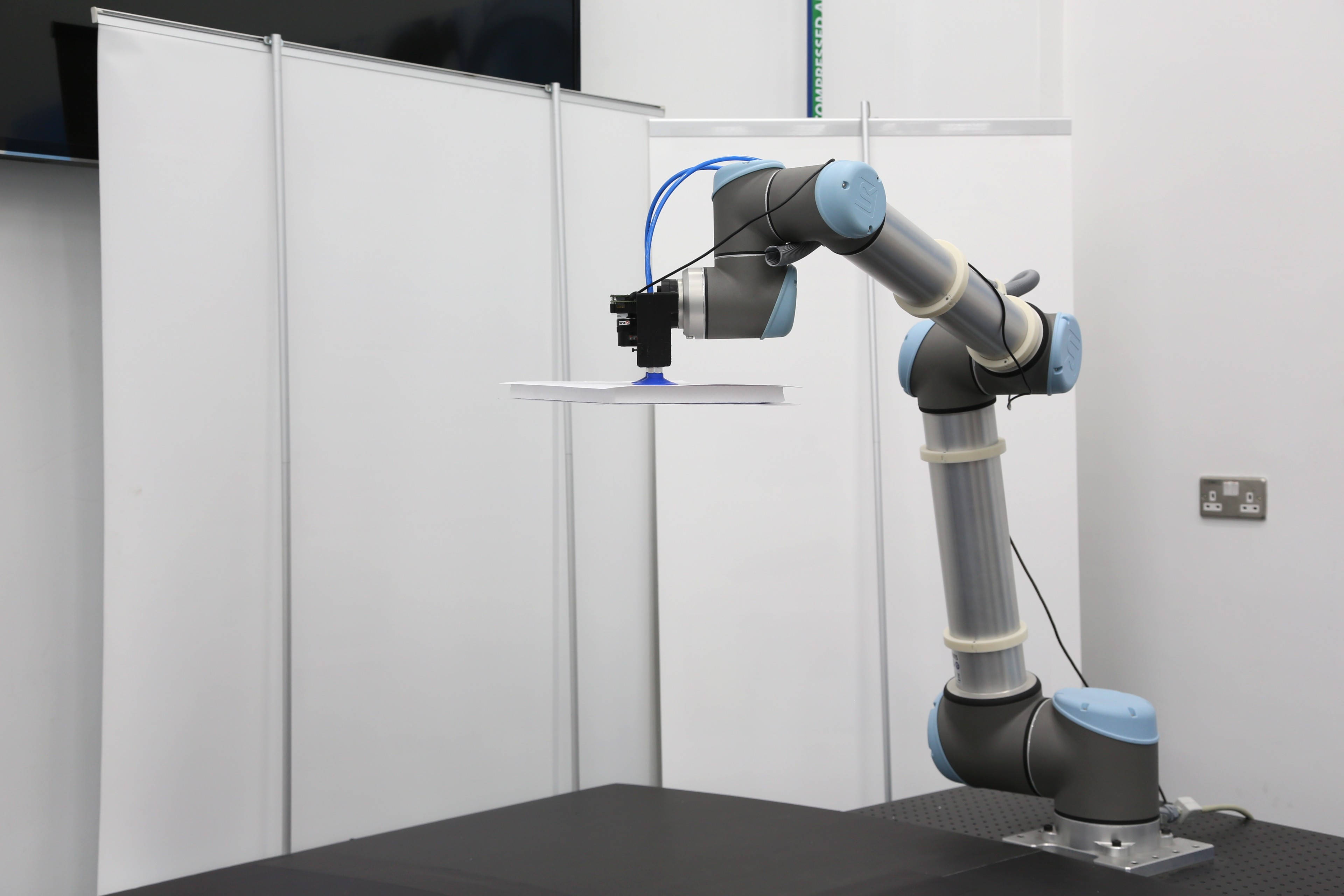}}\label{n}\hfil
\subfloat[Place]{\includegraphics[width=0.3\textwidth, height=0.21\textwidth]{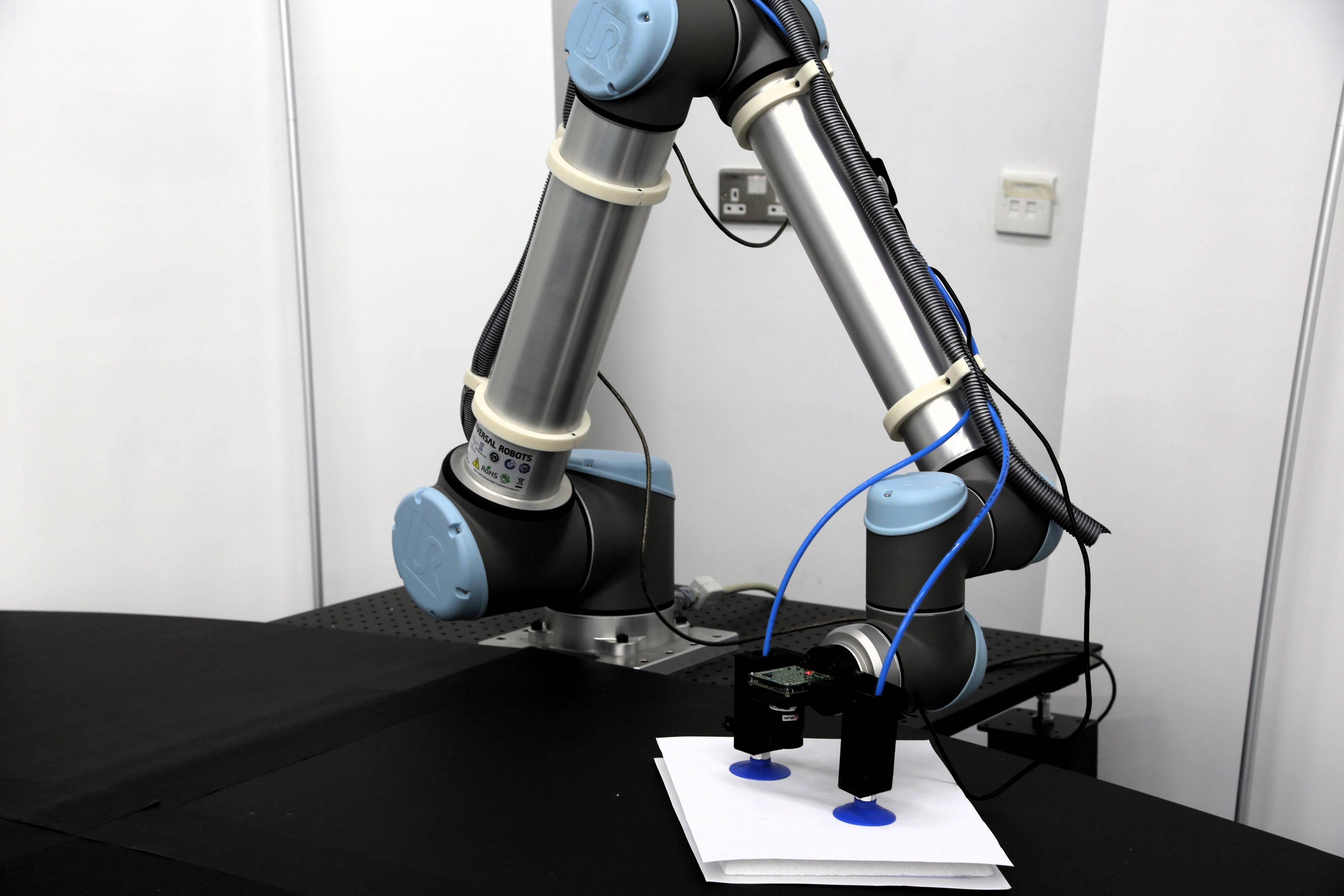}}\label{o}

\caption{Sequence of visual servoing operations to perform object manipulation task.}
\label{complete_manipulation}
\end{figure*}

\section{experimental validation of ebvs}

This section describes and discusses the results  of the experiments conducted to validate the proposed EBVS approach.

\subsection{Experimental Setup and Protocol}

The proposed method of visual servoing was incorporated in a top-down grasping paradigm to test its performance and applicability to real world smart manufacturing applications. The experimental setup consists of a Universal Robots UR10 6-DOF arm, a custom-made vacuum gripper, and a Dynamic and active pixel vision sensor (DAVIS240C) placed in an eye-in-hand configuration as displayed in Fig \ref{1_system_real_pic}. The DAVIS240C provides a spatial resolution of $240\times180$, a minimum latency of 12 microseconds, bandwidth of 12 MEvent/second and a dynamic range of 120 dB.

To successfully pick and place an object, the end-effector is first driven into alignment with the target object using the process described in section \ref{Method}. During the exploration and reaching phase, the end-effector's movement is constrained to a 2D plane perpendicular to the camera's optic axis. Once the end-effector is aligned with the target object, the end-effector translates in the camera's optic axis direction until contact with the object is achieved. Subsequently, the vacuum grippers are activated to grasp the target and relocate it to a desired location. Given limitations of the UR10's reach and the camera's field of view, the workspace of the experiments was limited to a 1.2 x 1.0 m virtual rectangle in front of the robotic platform. 

To evaluate EBVS performance against different geometries, experiments were carried out with three different objects; a triangular prism, a cuboid and a pentagonal prism. 

\subsection{Experimental Results}

Fig. \ref{complete_manipulation} shows the various stages of the proposed EBVS method for a visual servoing trial with a cuboid. For each stage, the robotic platform is displayed along with the corresponding heat-map of corner events and SAVE. During the exploration phase, the end-effector first moves towards a random virtual event \(p_{vr}\) to trigger events in the scene and update the heat-map. Based on the heat-map, the EBVS algorithm detects the object's high level features. Once contiguity is achieved in these features, the robot switches to the reaching phase where it moves towards the object's centroid \(p_{voc}\). The robot then enters the alignment phase where the grippers are rotated to achieve a stable grasp. Finally, the robot enters the grasping and manipulation phase to pick the object and place it in a desired location. By comparing the heatmaps and the SAVE with the top view pictures, the accuracy of the corner detection and tracking approach is demonstrated. Consequently, the centroid of the object in the SAVE is correctly inferred. As such, the proposed EBVS approach successfully drives and aligns the end-effector with the object prior to initiating the grasp. 

\begin{figure}[h!]
    \centering
    \includegraphics[width=0.45\textwidth]{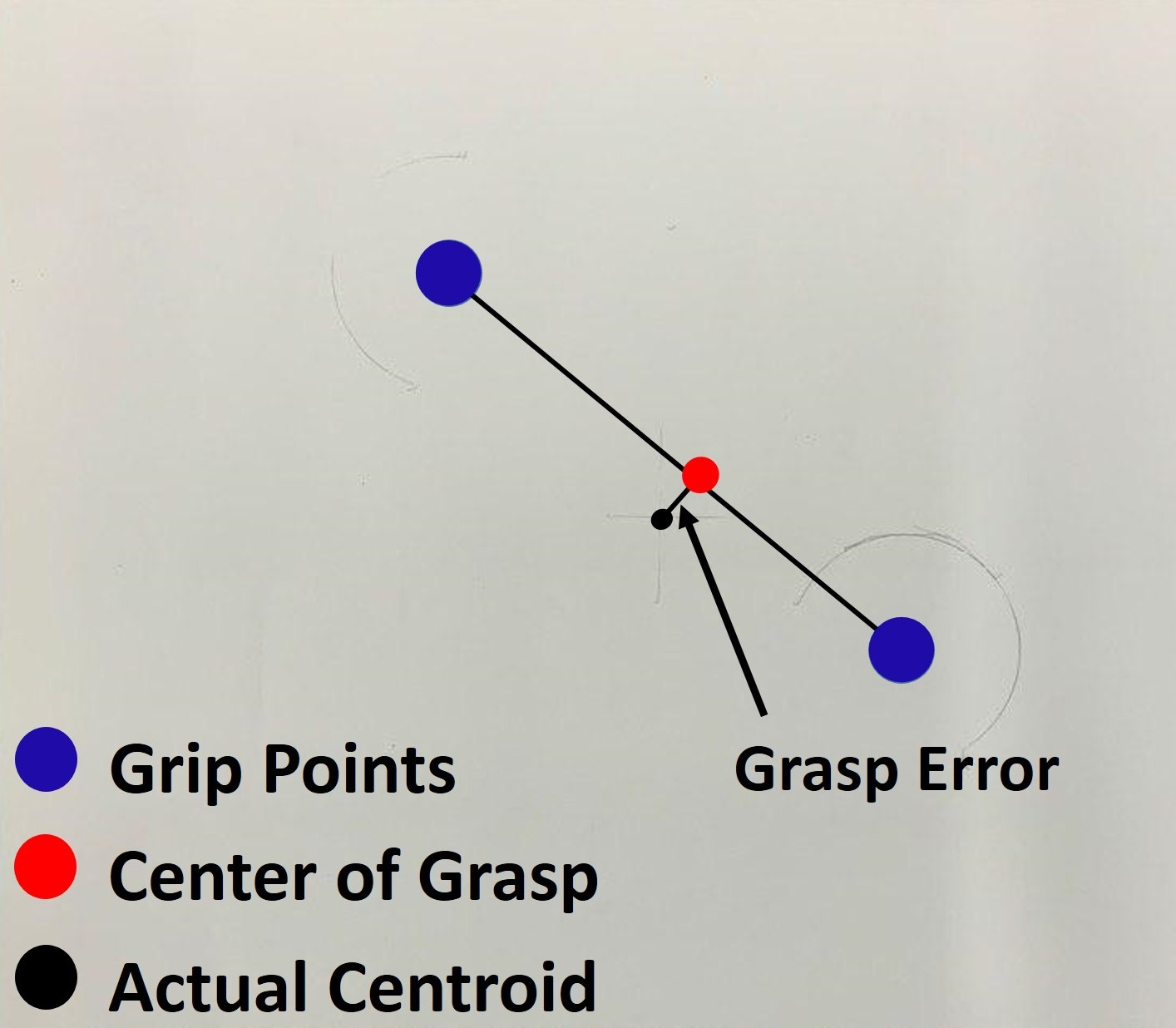}
    \caption[width=0.45\textwidth]{Grasp error}
    \label{fig:9_grasp_error}
\end{figure}

The same experiment in Fig. \ref{complete_manipulation} was repeated five times with a different placement of the object in the workspace. Table. \ref{tab:EXP_cuboid_results} shows the results of these experiments in terms of the grasp errors \(e_{grasp}\) and the number of times tracking was lost and the algorithm switched back to detection mode \(N_{switch}\). The grasp error is defined as the distance between the center of the two gripping points and the true object's centroid as illustrated in Fig. \ref{fig:9_grasp_error}. 

\renewcommand{\arraystretch}{1.5}
\begin{table}[]
    \caption{EBVS results with a rectangular shape}
    \centering
    \begin{tabular}{c|c|c}
            \textbf{Trial} & \boldmath{\(e_{grasp}\)} \boldmath{(mm)} &
            \boldmath{\(N_{switch}\)} \\
        \hline
        1 & 14.0 & 0 \\
        \hline
        2 & 19.0 & 0 \\
        \hline
        3 & 5.0 & 1 \\
        \hline
        4 & 17.0 & 0 \\
        \hline
        5 & 15.0 & 1 \\
        \hline
        \hline
        Average & 14.0 & 0.4 \\
    \end{tabular}
    \label{tab:EXP_cuboid_results}
\end{table}

Experiments were also carried out with different object shapes. Table \ref{tab:EXP_shapes_results} shows experimental results across five trials for three different geometrical shapes.

\renewcommand{\arraystretch}{1.5}
\begin{table}[]
    \caption{EBVS results with different geometries}
    \centering
    \begin{tabular}{c|c|c|c|c}
            \textbf{Shape}  & 
            \textbf{Trials} & \begin{tabular}{@{}c@{}}\boldmath{Mean } \\ \boldmath{\(e_{grasp}\)} \boldmath{(mm)}\end{tabular} &
            \begin{tabular}{@{}c@{}}\boldmath{Maximum} \\ \boldmath{\(e_{grasp}\)} \boldmath{(mm)}\end{tabular} &
            \begin{tabular}{@{}c@{}}\boldmath{Maximum} \\ \boldmath{\(N_{switch}\)}\end{tabular} \\
        \hline
        Triangle & 5 & 10.2 &   16.0 & 1 \\
        \hline
        Rectangle & 5 & 14.0 &  19.0 & 1\\
        \hline
        Pentagon & 5 & 24.2 &  45.5 & 4\\
        \hline
        \hline
        All Shapes & 5 &  16.1 & 45.0 & 4 \\
    \end{tabular}
    \label{tab:EXP_shapes_results}
\end{table}

In all the experiments, the presented visual servoing approach was capable of successfully tracking and grasping the target object with both vacuum grippers adhering to the object. The average grasp error for all the experiments was 16.1mm. These errors are mainly attributed to design imperfections such as the misalignment of the camera optic axis with the workspace plane and the skewed positioning of the cemera with respect to the center of the vacuum grippers. Enhancing the proposed method to such irregularities would be the one objective for future studies.

\begin{figure}[h!]
    \centering
    \includegraphics[width=0.45\textwidth]{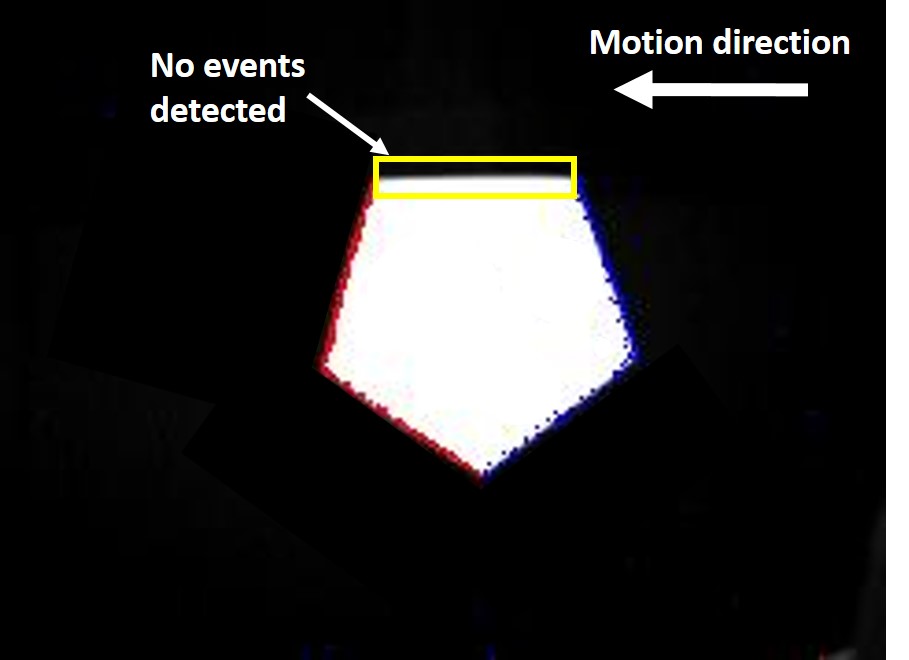}
    \caption[width=0.45\textwidth]{Failure to trigger events when moving parallel to an edge}
    \label{fig:10_parallel_events}
\end{figure}

The conducted experiments show that the proposed algorithm loses track more often with the pentagon shape; this in turns affects the accuracy of grasping as a larger deviation from the true object centroid was observed. As shown in Fig. \ref{fig:10_parallel_events}, when the neuromorphic camera moves parallel to an edge, it is less-likely to trigger events corresponding to this edge. As a result, the event-based harris corner detection fails to detect the corners associated with edges parallel to the camera's movement; causing EBVS to lose track. As a pentagon shape has edges with more varied slopes than a rectangle or a triangle, it is a more probable case for EBVS to encounter this shortcoming. A possible solution would be a filtering mechanism that determines the most reliable corners for EBVS tracking based on the camera's velocity vector. Such modifications would be the focus of further development to EBVS, and can be highly beneficial to other event-based visual tracking applications. 

\section{Conclusion and Future Work}

This study introduces a purely event-based visual servoing method that detects and tracks high-level features in a scene to perform a pick and place task suitable for smart manufacturing applications. A detailed explanation of the novel multi-stage servoing approach is presented, where three layers of active events are devised to process the incoming stream of events. Based on these layers, the gripper is accurately driven towards and aligned with the target object for grasping and placement.

Experiments validate the proposed EBVS method for use with objects of different geometrical features without the need for re-tuning or adaptation. The platform was able to precisely grasp objects placed randomly in the workspace with a 100\% success rate. For future work, we plan to improve the performance of the presented procedure by accounting for alignment uncertainties and augmenting an optimal motion planning scheme.


\bibliographystyle{IEEEtran}
\bibliography{visual_servoing_event_based}  

\begin{thebibliography}{10}
\providecommand{\url}[1]{#1}
\csname url@samestyle\endcsname
\providecommand{\newblock}{\relax}
\providecommand{\bibinfo}[2]{#2}
\providecommand{\BIBentrySTDinterwordspacing}{\spaceskip=0pt\relax}
\providecommand{\BIBentryALTinterwordstretchfactor}{4}
\providecommand{\BIBentryALTinterwordspacing}{\spaceskip=\fontdimen2\font plus
\BIBentryALTinterwordstretchfactor\fontdimen3\font minus
  \fontdimen4\font\relax}
\providecommand{\BIBforeignlanguage}[2]{{%
\expandafter\ifx\csname l@#1\endcsname\relax
\typeout{** WARNING: IEEEtran.bst: No hyphenation pattern has been}%
\typeout{** loaded for the language `#1'. Using the pattern for}%
\typeout{** the default language instead.}%
\else
\language=\csname l@#1\endcsname
\fi
#2}}
\providecommand{\BIBdecl}{\relax}
\BIBdecl

\bibitem{chaumette2006visual}
F.~Chaumette and S.~Hutchinson, ``Visual servo control. i. basic approaches,''
  \emph{IEEE Robotics \& Automation Magazine}, vol.~13, no.~4, pp. 82--90,
  2006.

\bibitem{hutchinson1996tutorial}
S.~Hutchinson, G.~D. Hager, and P.~I. Corke, ``A tutorial on visual servo
  control,'' \emph{IEEE transactions on robotics and automation}, vol.~12,
  no.~5, pp. 651--670, 1996.

\bibitem{vanarse2016review}
A.~Vanarse, A.~Osseiran, and A.~Rassau, ``A review of current neuromorphic
  approaches for vision, auditory, and olfactory sensors,'' \emph{Frontiers in
  neuroscience}, vol.~10, p. 115, 2016.

\bibitem{indiveri2000neuromorphic}
G.~Indiveri and R.~Douglas, ``Neuromorphic vision sensors,'' \emph{Science},
  vol. 288, no. 5469, pp. 1189--1190, 2000.

\bibitem{kragic2002survey}
D.~Kragic, H.~I. Christensen \emph{et~al.}, ``Survey on visual servoing for
  manipulation,'' \emph{Computational Vision and Active Perception Laboratory,
  Fiskartorpsv}, vol.~15, p. 2002, 2002.

\bibitem{cui2020visual}
L.~Cui, H.~Wang, X.~Liang, J.~Wang, and W.~Chen, ``Visual servoing of a
  flexible aerial refueling boom with an eye-in-hand camera,'' \emph{IEEE
  Transactions on Systems, Man, and Cybernetics: Systems}, 2020.

\bibitem{wang2016visual}
H.~Wang, B.~Yang, Y.~Liu, W.~Chen, X.~Liang, and R.~Pfeifer, ``Visual servoing
  of soft robot manipulator in constrained environments with an adaptive
  controller,'' \emph{IEEE/ASME Transactions on Mechatronics}, vol.~22, no.~1,
  pp. 41--50, 2016.

\bibitem{wang2020eye}
X.~Wang, G.~Fang, K.~Wang, X.~Xie, K.-H. Lee, J.~D.-L. Ho, W.~L. Tang, J.~Lam,
  and K.-W. Kwok, ``Eye-in-hand visual servoing enhanced with sparse strain
  measurement for soft continuum robots,'' \emph{IEEE Robotics and Automation
  Letters}, 2020.

\bibitem{gallego2019event}
G.~Gallego, T.~Delbruck, G.~Orchard, C.~Bartolozzi, B.~Taba, A.~Censi,
  S.~Leutenegger, A.~Davison, J.~Conradt, K.~Daniilidis \emph{et~al.},
  ``Event-based vision: A survey,'' \emph{arXiv preprint arXiv:1904.08405},
  2019.

\bibitem{vasco2016fast}
V.~Vasco, A.~Glover, and C.~Bartolozzi, ``Fast event-based harris corner
  detection exploiting the advantages of event-driven cameras,'' in \emph{2016
  IEEE/RSJ International Conference on Intelligent Robots and Systems
  (IROS)}.\hskip 1em plus 0.5em minus 0.4em\relax IEEE, 2016, pp. 4144--4149.

\bibitem{mueggler2017fast}
E.~Mueggler, C.~Bartolozzi, and D.~Scaramuzza, ``Fast event-based corner
  detection,'' in \emph{in 28th British Machine Vision Conference
  (BMVC)}.\hskip 1em plus 0.5em minus 0.4em\relax University of Zurich, 2017.

\end{thebibliography}

\end{document}